\documentclass[11pt]{article}

\usepackage[preprint]{acl}

\usepackage[T1]{fontenc}
\usepackage{mathptmx}
\usepackage{helvet}
\usepackage{courier}
\usepackage{url}
\usepackage{booktabs}   
\usepackage{amsmath}    
\usepackage{graphicx}
\usepackage{pgfplots}   
\pgfplotsset{compat=1.18}
\usepackage{colortbl}   
\usepackage{xcolor}
\usepackage{tabularx}   

\usepackage{tcolorbox}
\tcbuselibrary{skins,breakable}

\definecolor{failbg}{gray}{0.90}
\definecolor{passbg}{gray}{0.97}
\newcommand{\vFail}{\cellcolor{failbg}\textbf{FAIL}}
\newcommand{\vPass}{\cellcolor{passbg}pass}
\newcommand{\vNC}{\cellcolor{passbg}not computable}

\definecolor{takeawayaccent}{HTML}{44607A}  
\definecolor{takeawaytint}{HTML}{F3F6F9}    

\newcounter{takeaway}

\newtcolorbox{takeawaybox}[1]{%
  enhanced, breakable,
  colback=takeawaytint,
  colframe=takeawayaccent,
  colbacktitle=takeawayaccent,
  coltitle=white,
  boxrule=0.6pt,
  rounded corners,
  arc=2.5pt,
  title={\ensuremath{\triangleright}~\textbf{Takeaway \thetakeaway: #1}},
  fonttitle=\small,
  toptitle=1.5pt, bottomtitle=1.5pt,
  left=7pt, right=7pt, top=5pt, bottom=6pt,
  before skip=0.9em, after skip=0.9em,
  fontupper=\small,
}

\newcommand{\takeaway}[2]{%
  \par
  \refstepcounter{takeaway}%
  \begin{takeawaybox}{#1}%
    #2%
  \end{takeawaybox}%
  \par
}

\makeatletter
\renewcommand{\@seccntformat}[1]{%
  \csname #1prefix\endcsname
  \csname the#1\endcsname
  \csname #1suffix\endcsname
  \quad
}
\makeatother

\makeatletter
\newcommand{\RequireGenerated}[1]{%
  \@ifundefined{#1}{%
    \GenericError{}{%
      PROVENANCE ERROR: macro #1 is not defined%
    }{}{%
      This macro must come from generated/results_macros.tex, which is written
      by: python training/holdout_eval.py --writeup <model dirs> --verify-against
      <run json>. Re-run the emitter. Do NOT define it by hand -- a number
      typed here is a number nothing computed.%
    }%
  }{}%
}
\makeatother

\newcommand{\ResultFingerprint}{a288e8ca47f375f6}

\newcommand{\CorpusRows}{256}
\newcommand{\CorpusFlagged}{29}
\newcommand{\CorpusHeld}{23}
\newcommand{\CorpusFlaggedAndHeld}{19}
\newcommand{\CorpusHeldOnly}{4}
\newcommand{\CorpusFlaggedOnly}{10}
\newcommand{\CorpusInapplicable}{50}
\newcommand{\CorpusUnauthored}{19}
\newcommand{\CorpusSecondPhrasings}{1}
\newcommand{\CorpusInventory}{165}
\newcommand{\CorpusFlaggedInInventory}{8}
\newcommand{\SeedFloor}{3,000}
\newcommand{\SeedRowsPerSeed}{50}
\newcommand{\SeedOurs}{165}
\newcommand{\SeedOursRaw}{165}
\newcommand{\SeedFloorRaw}{3000}

\RequireGenerated{ResultFingerprint}

\title{\textbf{KinyaMed: Seeds, Not Rows}\\
\large What a Corpus Requirement Written in the Wrong Unit Fails to Constrain%
\thanks{Build fingerprint \texttt{\ResultFingerprint}. Every number in this paper
re-derives from a clean clone of the repository with \texttt{make reproduce},
except those marked \textbf{NOT REPRODUCIBLE} where they appear: a small number
of figures rest on artefacts that were never committed.}}
\author{Marius Bayizere\\
  Independent Researcher, Kigali, Rwanda\\
  \texttt{bayizeremarius119@gmail.com}\\
  \url{https://github.com/mariusbayizere/healthguard-ai}\\
  {\small The system is named KinyaMed; the repository retains its original
  name, \texttt{healthguard-ai}. They are the same artefact.}}
\date{\today}

\begin{document}
\maketitle

%
%

\begin{abstract}
Triage decides who is seen first. Building an urgency classifier for
patient-voice Kinyarwanda, we found our specification could be met without
producing anything it was meant to secure. We report that, and the instruments
that detect it, instead of a classifier. Designed for the four languages a Rwandan health
centre receives, with every instrument per-language: sentences are authored in
all four arms and rows generate in one, because the frame slots those three
need do not exist. Our requirement asked for one million
examples; generation produced them in 130 seconds. It fails four of the nine
quality gates in that specification, six when rows are attributed to their
source sentence. The binding gate counts distinct authored
seed phrases, not rows: at our 165, no corpus passes at any row count. Two shortfalls follow and differ: 2{,}835 further sentences to pass the
seed-count gate, 19{,}835 to reach the stated million rows, because a separate
gate caps a seed at 50 rows. Rows come from a machine at
7{,}700 per second; seeds from clinicians.
A row count constrains the cheap quantity, leaves the expensive one free, and
so does not constrain quality at all.

Two further negatives follow. An evaluation set of 17{,}942 rows built from
nine distinct sentences supports no verdict: our gate, which counts distinct
sentences, refuses 38 of its cells and reports nothing. A model trained on a
corpus of uniform surface form changes its predicted urgency for 31.5\% of
inputs under capitalisation and 21.0\% under a single typo, reported as
measurement and not attribution. A model directory shipped without its
tokenizer loads without error and answers its class prior on input it cannot
read, with well-formed probabilities. No figure here is evidence of model
quality; the contribution is the apparatus and the negative results it
produced, reproducible from a clean clone except where marked \textbf{NOT
REPRODUCIBLE}.
\end{abstract}

\begin{center}
\begin{minipage}{0.95\linewidth}
\small\textbf{Keywords:} low-resource NLP, Kinyarwanda, clinical triage, dataset
quality, negative results, evaluation methodology, reproducibility.
\end{minipage}
\end{center}

\vspace{1em}

%
%

\section{Introduction}
\label{sec:introduction}

A nurse facing a full waiting room has to decide who is seen first. We set out to
build an urgency classifier for patient-voice Kinyarwanda to support that
decision, and we report instead what we found while trying to build one honestly:
a specification that could be satisfied without producing anything it was meant
to secure, an evaluation set that could not support a verdict, and two
\emph{silent} failure modes, a tokenizer that can be left behind and a
classifier whose answer moves under surface variation, both of which leave every
signal a practitioner would check looking healthy.

This is a negative-results and methodology paper. Model numbers appear in it,
because several of the failures we report are failures about those numbers, but
\textbf{none is offered as evidence of model quality}; Section~\ref{sec:res:none}
states the distinction and the reasons for it.

The system is designed for the four languages a Rwandan health centre receives,
Kinyarwanda, English, French and Kiswahili, and the design is per-language
throughout: the corpus gates count seeds per language, the power derivation fixes
a minimum per language crossed with urgency class, and the person and relation
rulings are recorded per arm. \textbf{The corpus that exists is Kinyarwanda
only}: sentences are authored in all four arms and rows generate in one, because
the openers, contexts and closers a row is wrapped in do not exist for English,
French or Kiswahili (Table~\ref{tab:arms}).

\paragraph{A corpus requirement in rows constrains the cheap quantity.} Our
functional requirement asked for one million examples. Generating them took 130
seconds. Run against the nine corpus-quality gates written in the same
specification, that corpus fails four as generated, and six where its rows can be
attributed to a source sentence. The gate that binds counts \emph{distinct
authored seed phrases}, and our generator has 165: at that inventory the largest
corpus passing every gate is zero rows, whatever the row count. Two distances
follow and they are not the same number. Passing the seed-count gate needs
2{,}835 further authored sentences; reaching the one million rows the
requirement names needs 19{,}835, because a separate gate caps any one seed at
50 rows. Neither distance is measured in rows, and the missing 999{,}000 rows
are not the shortfall that matters. We take the
general form of this seriously, because it is not specific to our generator:
expansion is the cheap operation in any templated or model-expanded corpus, so a
requirement written in rows leaves the expensive quantity, independently
authored source items, entirely unconstrained.

\paragraph{The language problem is not a translation problem.} Our first corpus
was generated by a non-speaker from English clinical concepts and machine
translation, and it was not usable. A Kinyarwanda speaker reading it found
grammatical sentences that no patient would say: English syntax carried across
intact, register drifted toward the clinical, and borrowings that real speech
uses freely were avoided in favour of ``purer'' constructions. That is not a
quality gap a larger model closes. It is the difference between text about a
language and text in it, and the only remedy we found was to have the phrases
authored rather than translated, which is the same constraint the corpus gates
express, arrived at from the other direction.

\paragraph{Small corpora make evaluation harder, not easier.} A corpus assembled
from a few hundred authored sentences and combinatorially expanded looks large
and is not. Our evaluation set contains 17{,}942 rows built from nine distinct
sentences. Our deployment gate counts distinct sentences rather than rows, and on
that set it refuses 38 of its cells and reports nothing. We state the
distinct-sentence count beside every figure in this paper, and never the row
count alone.

\paragraph{Two process failures worth more than the model.} These are a
different pair from the two silent failure modes above, and the distinction
matters: those are properties of the artefact, these are failures of our own
reading. Both were recorded correctly and acted on late.

The first is an evaluation split that never worked. Its manifest recorded that
100\% of its evaluation rows were built on phrases also present in training:
in a machine-readable field, in the commit that created the pipeline. The two
fields anyone read were both honestly zero. The measurement was not missing; the
reading was.

The second is our acceptance gate. It was a single condition, recall on the
critical class, chosen for the defensible reason that missing a critical
presentation is the failure that matters. A configuration passed it while
assigning the critical label to almost every urgent case, and its precision on
that class sat \emph{below} what merging the two classes earns for free on the
same rows. The gate certified the worse of two configurations and had no term in
which over-triage could be expressed.

\paragraph{Scope, stated first rather than last.} This is a single-author
software project with no clinical collaborator to date. The corpus is Kinyarwanda
only. No clinician has validated any label, no second speaker has reviewed the
corpus, and no document in the repository authorises assigning urgency from a
written report about a patient nobody has examined
(Section~\ref{sec:related}). Nothing here has been used with any patient. The
trained artefact that exists is an audit artefact: we use it to exercise the
serving path and the gate's refusal, we do not compare against it, and we do not
plan to retrain it. Section~\ref{sec:limitations} is the section we would most
want a reader to weigh.

\paragraph{We do not claim priority.} We are not aware of a prior patient-voice
medical triage dataset for Kinyarwanda, and the resource audit described in
Section~\ref{sec:related} did not locate one. We have not run a systematic search
of the field, and we make no stronger claim than that.

\paragraph{Contributions.} Our audit yields five findings.

\textbf{(1) A corpus requirement written in rows does not constrain quality.}
The binding quantity is distinct authored seed phrases, not rows, and the two
are separated by a factor a generator supplies for free. At our inventory of
165 seeds, no corpus passes the gates at any row count, and the distance to the
specification is 2{,}835 sentences to pass the seed gate or 19{,}835 to reach
the stated million rows, never 999{,}000 rows.

\textbf{(2) Cluster structure, not row count, decides what a held-out set can
show.} An evaluation set of 17{,}942 rows built from nine distinct sentences
supports no verdict. We derive what a set must contain before any gate verdict
is possible, and our gate refuses on the set that exists rather than reporting a
number it cannot support.

\textbf{(3) A single-metric safety gate certified a model that had abandoned an
urgency class.} The gate had no term in which over-triage could be expressed.
The replacement is three conditions, of which one is derived exactly, one is
inherited without a verified source, and one rests on a margin we do not derive,
and we label which is which.

\textbf{(4) Two failure modes leave every output-side signal healthy.} A model
directory shipped without its tokenizer, and a classifier whose answer moves
under surface variation. Neither is visible from probabilities, accuracy or
label order; each was caught by a property check against something outside the
model.

\textbf{(5) A recurring failure we name \emph{recorded but unread}.} Thirteen
times a fact was recorded correctly, in a machine-readable place, and went unread
until something unrelated forced a look. Six are one shape: a leakage field, a
duplicated threshold, three emitted tables no file included, a docstring
disagreeing with its own constant, a gate that could not fail, and a required
gloss field that is a placeholder in 13 of the 15 phrases our figures rest on.
Four are instruments that reported on what they had not read, among them a
checksum that confirmed an incomplete archive was identical to itself. Two are a
count or a sentence correct when written and never re-derived after the thing it
described changed. The last is a tool that read correctly and was taken to have
answered a wider question than it was asked, and it was introduced by a fix that
was right about its own subject. None is a measurement failure. Each is a fact
available where the thing that needed it does not look
(Section~\ref{sec:disc:unread}).

The apparatus that produces all five re-derives from a clean clone with one
command, except where marked \textbf{NOT REPRODUCIBLE}.

\begin{center}
\fbox{\parbox{\dimexpr\linewidth-2\fboxsep-2\fboxrule\relax}{\small\raggedright
\textbf{Provenance of every number.} No number in this paper is offered as
evidence of model quality.
Each number in it is a property of a corpus, a specification, or an artefact's
behaviour under perturbation, and \path{reports/PAPER_NUMBERS.md} lists every
one with the committed script or record that produces it. Figures whose script
was never committed are marked NOT REPRODUCIBLE where they appear, or cut. The
gate-degeneracy figures are generated macros written by
\path{training/holdout_eval.py} \texttt{-{}-writeup}; they carry the fingerprint of the
weights, the evaluation data and the commit that produced them.
}}
\end{center}

%

\section{Related work}
\label{sec:related}

\subsection{Kinyarwanda in current African-language model work}
\label{sec:related:afriquellm}

Kinyarwanda is represented in the current reference point for continued
pre-training on African languages. \citet{yu2026afriquellm} adapt five base
models to twenty African languages and report their per-language data mixture;
their Table~1 gives Kinyarwanda 481M raw tokens, upsampled over two epochs to
1.07B, with 13M synthetic tokens added. That is the scale at which general
Kinyarwanda text is available, and it stands in a useful contrast to the
quantity this paper is about. Our binding constraint is not tokens of running
text but \SeedFloor{} independently authored clinical sentences in the patient's
own voice, a quantity no crawl supplies and no upsampling factor changes.

\subsection{Why we make no comparison with a triage instrument}
\label{sec:related:instruments}

We compare against no established triage instrument, and the reason is not
licensing. It is that every instrument we examined is defined on a different kind
of input from the one this system receives, which makes a comparison not merely
unavailable but ill-posed.

The instrument our own specification named as the basis for its three categories
is the WHO Emergency Triage Assessment and Treatment (ETAT) manual for
participants~\cite{who2005etat}. Reading it settled the question
against us, and the details are worth stating because they are specific:

\begin{itemize}
\item \textbf{It is defined on examination signs.} ``Triage is the process of
  rapidly \emph{examining} all sick children when they first arrive in
  hospital''; ``Triage of patients involves \emph{looking for signs} of serious
  illness or injury'' (Module One, pp.~3--4). Each emergency sign is elicited by
  hand or eye: taking the child's hand to judge warmth, pressing the nail bed and
  timing capillary refill, pinching the abdominal skin, comparing the child's
  palms with the assessor's own (Modules Three to Five, pp.~26, 35--36, 44).
\item \textbf{It rules explicitly against history where history and observation
  could conflict.} On convulsions: ``This assessment depends on your observation
  of the child and not on the history from the parent \dots\ The child must be
  seen to have a convulsion during the triage process or while waiting in the
  outpatient department'' (p.~7). Where history is used (whether the child has
  diarrhoea, a history of choking, a referral note), it is asked of the
  caretaker in person, as an adjunct to an examination that still takes place.
\item \textbf{It does not address remote or written assessment.} Triage happens
  ``as soon as a sick child arrives in the hospital, well before any
  administrative procedure such as registration'' (p.~5). Across all 83 pages of
  the extracted text, the words \emph{telephone}, \emph{phone}, \emph{radio},
  \emph{remote}, \emph{SMS} and \emph{mobile} do not occur once.
\item \textbf{Its population is children.} ``a process of rapid triage for all
  children presenting to hospital'' (p.~1); adults appear only as equipment sizes
  (pp.~21, 46, 65). No upper age limit is stated anywhere in the document.
\item \textbf{Its categories are its own}: EMERGENCY, PRIORITY and NON-URGENT
  (p.~4; Chart~2, pp.~67--68), not the critical/urgent/routine triple our
  specification uses, and not the Emergency Severity Index levels our
  specification also cites in the same glossary entry.
\end{itemize}

The point generalises beyond ETAT. Bedside triage instruments share the property
that matters here: a trained assessor observes a patient who is present. Our
system receives text written by a patient or a relative who has not been
examined, often before any contact with a health worker. An instrument defined on
observed signs cannot be a reference standard for a system that never observes
anything, and a number comparing the two would measure agreement between
incommensurable procedures.

\paragraph{The absence is the finding.} Validated triage instruments exist, and
we do not claim otherwise. Telephone triage protocols are in routine clinical
use, nurse-led telephone triage has been studied for decades, and symptom
checkers have been audited against clinical vignettes. What we could not locate,
in any language, is an instrument validated for the specific task this system
performs: assigning urgency from a patient's own \emph{written} free text, in the
patient's own words, without examination and without a clinician in the loop
asking follow-up questions. The difference is not a technicality. Telephone
triage is interactive, and symptom-checker audits are typically run against
vignettes written by clinicians rather than text written by patients. If such an
instrument exists, it is the reference standard this
work needs and we did not find it; if one does not, then the construct this
system would measure has not been established by anyone, and that is the more
important statement. We record it as an open question addressed to clinical
readers rather than as a gap we have closed, and we treat every urgency label in
our corpus as an unvalidated construct until a clinician rules otherwise.

\subsection{Clinical guidance consulted during construction}

Where published guidance existed for a presentation, it was consulted while
concepts were written, and no text was reproduced from any of it: a clinical fact
may be cited, the wording a document expresses it in is not ours to copy. We name
no document here and cite none, because the assignments that would justify naming
them have never been checked against the sources (Section~\ref{sec:method}). The
licensing analysis that governed the no-reproduction rule is recorded in the
repository.

\subsection{Kinyarwanda language resources}

Kinyarwanda NLP has usable encoders and retrievers and very little else, and
almost nothing in a patient's voice. The resources we assessed fall into three
groups.

\paragraph{News-domain corpora.} KINNEWS and KIRNEWS~\cite{niyongabo2020kinnews} provide 3{,}449 Kinyarwanda
news articles collected from a single news site. MasakhaNER~\cite{adelani2021masakhaner}
and MasakhaNER 2.0~\cite{adelani2022masakhaner2} provide named-entity-annotated
Kinyarwanda and Swahili text, also news. Both are useful
for pretraining and neither contains clinical language or patient speech. We use
neither: MasakhaNER is non-commercial share-alike, and the KINNEWS licence is not
stated in its repository at all, which we treat as unusable until confirmed
rather than as permissive by default.

\paragraph{Patient-facing health text.} We assessed further candidate resources
and use none of them. Two that we measured directly are omitted here rather than
described: we could not establish a citable identity for either, and describing a
source we cannot name would be inconsistent in a paper about provenance. The
assessments are in the repository.

\paragraph{Community health worker questions.} The most useful Kinyarwanda
clinical resource we found is a set of real community health worker questions with
Kinyarwanda originals and clinician answers, a subset of which is released
alongside a study of large language models for frontline healthcare support in
low-resource settings~\cite{rutunda2026llm}. \textbf{We quote no counts and no
licence for it.} The deposited item returned an HTTP error to every automated
request during verification, so the size of the released subset, the size of the
full collection and the licence terms are recorded in our own audit notes and were
never read from the item itself. What we can say without the item is what the
study states: the released portion is a subset, and an access route rather than a
download governs the rest.

That resource is close to ours in setting and different in kind. It records what
a health worker asks a clinician; we need what a patient says to a health worker.
It is a genuine sample of real language, which our corpus is not, and its
existence is the strongest argument that the authored approach we took is a
stopgap rather than a destination.

\subsection{Where this work sits}

We are not aware of a prior patient-voice medical triage dataset for
Kinyarwanda, and the audit above did not locate one. That is a statement about
our search, not about the field. What we can say more firmly is narrower and
more useful: the resources that exist are news-domain or institutional, the one
clinical Kinyarwanda corpus we found is in a health worker's voice rather than a
patient's, and neither the concept taxonomy nor the licensing position we
adopted could be inherited from any of them.

\section{Method}
\label{sec:method}

\subsection{Corpus v1, and why it was replaced}

The first corpus was generated, not authored. A template generator combined a
small inventory of symptom phrases with frame slots (an opener, a subject, an
onset, a context and a closer) to produce 1{,}000{,}000 rows across four
languages (Kinyarwanda, English, French, Swahili) from 184 template-drafted
phrases. The phrases themselves were written by a non-speaker working from
English clinical concepts.

Two independent findings retired it.

\paragraph{The language was not verifiable.} No Kinyarwanda speaker had read any
of it. When one did, the assessment was that the phrases were grammatical but
not what a patient would say: syntax calqued from English structure, register
drifted upward toward the clinical, borrowed words that speech uses freely
(\emph{malaria}, \emph{pressure}, \emph{sugar}) avoided in favour of ``pure''
constructions, and descriptions that were too complete for speech. The speaker
also rejected the noun-phrase form the generator required (phrases were
written to slot in after \emph{I have} / \emph{my wife has}), and rewrote the
existing phrases as full patient utterances. That change moved person out of the
frame and into the phrase, and it is the reason the v2 inventory carries a
first-person and a third-person row per concept.

\paragraph{One of the two evaluation splits never worked, and the manifest
recorded it from the first commit.} v1's family holdout was not a holdout and its
own frozen manifest said so, in a field nobody read; the manifest figures, the
separate cross-split contamination, and why v2 is \emph{worse} on that second
measure rather than better are in Appendix~\ref{app:v1leak}.

\subsection{Corpus v2}

v2 is monolingual Kinyarwanda: 330{,}000 rows generated from 165 distinct
phrases, drawn without replacement from each phrase's frame combinations. It is
a rebuild rather than a revision: a v1 result and a v2 result are not the same
experiment and do not belong in one table.

%
\begin{table*}[t]
\centering\small
\setlength{\tabcolsep}{4pt}

\textbf{Part A: one seed phrase and five of its 446 frame permutations.}

\begin{tabularx}{\textwidth}{r>{\raggedright\arraybackslash}X}
\toprule
& \textbf{Seed} (concept IF05, \emph{fever with generalised rash}, label URGENT) \\
& Mfite umuriro kandi mfite uduheri ku mubiri wose. \\
\midrule
1 & Nyabuneka, mfite umuriro kandi mfite uduheri ku mubiri wose kuva mu cyumweru gishize \\
2 & Muraho, mfite umuriro kandi mfite uduheri ku mubiri wose kuva hashize icyumweru nta miti imfasha. Murakoze. \\
3 & Mfite umuriro kandi mfite uduheri ku mubiri wose kuva mu gitondo kandi nta miti mfite. Urakoze. \\
4 & Mfite umuriro kandi mfite uduheri ku mubiri wose kuva mu cyumweru gishize. Abandi bana na bo bafite iki kibazo. Nkora iki? \\
5 & Nyabuneka, mfite umuriro kandi mfite uduheri ku mubiri wose kuva mu masaha abiri ashize. Byatangiye gitunguranye. \\
\bottomrule
\end{tabularx}

\vspace{0.8em}

\textbf{Part B: the same concept in two persons, and what no similarity rule catches.}

\begin{tabularx}{\textwidth}{l>{\raggedright\arraybackslash}Xr}
\toprule
Person & Phrasing & Characters \\
\midrule
first & Mfite umuriro kandi mfite uduheri ku mubiri wose. & 49 \\
third & \texttt{\{REL\}} afite umuriro n'uduheri ku mubiri wose. & 45 \\
\midrule
\multicolumn{2}{l}{Shared prefix} & \textbf{0} \\
\multicolumn{2}{l}{Longest common substring} & 23 \\
\multicolumn{2}{l}{Word Jaccard} & 4/10 = 0.400 \\
\bottomrule
\end{tabularx}
\\[2pt]{\footnotesize \texttt{\{REL\}} is the stored relation placeholder, expanded at generation time to one of 8 ruled relation terms; the counts above are computed from the stored form.}

\caption[A frame permutation, and what no similarity rule catches.]%
{What a frame permutation is, and what no similarity rule catches.
\textbf{Part A.} All 446 rows built on this seed carry the label
URGENT \emph{because the seed carries it}: the seed text is invariant and
only the opener, the time expression and the closing line move. They are one
observation, not
446, which is the cluster argument made visible rather than argued.
\textbf{Row 1 ends without terminal punctuation.} That is how the
generator emitted it and it is left untouched: it is a live instance of the
C9 surface-uniformity failure, visible in the example itself rather than only
in the gate's verdict.
\textbf{Part B.} The two phrasings share a prefix of 0 characters,
because the third person opens on the \texttt{\{REL\}} placeholder, and
their word
Jaccard of 0.400 is far below the 0.85 the near-duplicate gate
uses. Neither a prefix rule nor a similarity threshold would group them.
They are
in one phrase group
\emph{because the authoring record declares them one concept}, not because any
string comparison found them alike, and that is what makes this the case no
similarity rule catches.
Substituting a relation term for the placeholder would change the character
count and the shared prefix, which is why the counts are computed from the
stored form and the expansion is described in the note rather than shown as a
row.
\textbf{On the choice of IF05.} EX24 gives a cleaner Part B
(shared prefix 0, longest common substring 25 against IF05's 23) and is
not used because its \path{english_gloss} in the authoring record is a
placeholder rather than a gloss.
\textbf{Of the 15 distinct phrases in the reporting split,
2 carry a real English gloss
and they are the two persons of a single concept.}
\textbf{The remaining 13 phrases the reported figures rest
on are glossed only by a placeholder telling the author where to
look.}
That is a property of the
corpus rather than a note about which row was convenient, and it is why this
table could not be built from the cleaner pair.}
\label{tab:example}
\end{table*}

The table above is the argument of this paper in one page. Part A shows why a
row count says nothing: 446 rows, one observation. Part B shows why the leakage
guarantee cannot rest on a similarity rule, because the two phrasings of one
concept share no prefix at all and sit well below any duplicate threshold, and
are grouped only because the authoring record says they are one concept.

\subsection{Four language arms, one corpus}
\label{sec:method:arms}

The system is designed for four languages because a Rwandan health centre
receives all four. Every instrument in this paper is per-language: the corpus
gates count distinct seeds \emph{per language}, the power derivation fixes a
minimum \emph{per language crossed with urgency class}, and the person and
relation rulings are recorded per arm rather than assumed to transfer. The
corpus that exists is Kinyarwanda. Table~\ref{tab:arms} gives the inventory of
all four arms, emitted from the generator and the briefs rather than described.

%
\begin{table*}[t]
\centering
\small
\setlength{\tabcolsep}{4pt}
\begin{tabular}{>{\raggedright\arraybackslash}p{3cm}rrrrrr}
\toprule
Arm & Phrases & Drafts & Authored & Relations & Frames & Rows \\
\midrule
\textbf{Kinyarwanda} & 165 & -- & 2,281 & 8 & 43 (4/4) & 330,000 \\
English & 0 & 198 & 2,302 & 8 & 10 (1/4) & 0 \\
French & 0 & 194 & 2,301 & 8 & 10 (1/4) & 0 \\
Swahili & 0 & 0 & 2,300 & 7 & 10 (1/4) & 0 \\
\bottomrule
\end{tabular}
\caption{What each language arm contains. \textbf{Phrases} are authored by a
speaker of that arm; \textbf{drafts} are machine-written candidates no
speaker of that language has reviewed, counted separately because counting them
as phrases is the overstatement this paper is about. \textbf{Authored} counts
labelled sentences that exist and that the generator cannot use; no arm is zero
in it, so no row here can be read as an arm nobody has written yet.
\textbf{Frames} counts
openers, onsets, contexts and closers, with the number of those four slots that
are non-empty: a row needs all four, which is why three arms generate nothing
whatever their other columns say. Provenance differs within a column
(English's are wording from the v1 machine drafts; Swahili's are speaker authored).}
\label{tab:arms}
\end{table*}

\textbf{English, French and Kiswahili now demonstrate that claim rather than
illustrating it.} All three hold authored sentences, 2{,}302, 2{,}301 and
2{,}300 respectively, and all three generate \textbf{zero rows}. The sentences
exist; the corpus does not. What is missing is not authoring but the wrappers a
row is built from: each arm has \textbf{one of the four frame slots}, the onsets
left over from v1's machine drafts, and lacks openers, contexts and closers.
\textbf{The distance to a second generating arm is three frame slots, not a
single further sentence.} That is the seeds-not-rows argument turned around: we
have the expensive quantity in all four languages and cannot emit a row in three
of them, because the cheap quantity is absent.

\textbf{With the Kiswahili arm authored, one reading of this table closes.}
While that arm stood at zero in every column it could be read as work not yet
done, and the argument rested on two cases. It now rests on three, and no arm in
Table~\ref{tab:arms} is empty of authored sentences: \textbf{9{,}184 sentences
across four languages produce rows in exactly one}. The column that separates
that one from the rest is not \textbf{authored}, where the four arms sit within
twenty-one sentences of each other, but \textbf{frames}, where one arm has four
slots and three have one.

The column that decides the others is \textbf{frames}. A row is a phrase wrapped
in an opener, a time expression, a context and a closing line, and an arm
missing any of the four emits nothing whatever its phrase count. Only
Kinyarwanda has all four. The ten fragments each other arm carries are onsets
left over from v1's machine drafts, which is why three arms sit at zero rows and
why the phrase inventory is not the binding constraint on a four-language
corpus. The Kiswahili arm makes this hardest to argue against, because its brief
shipped deliberately without a Kiswahili draft so that a fluent draft could not
anchor the speaker's phrasing, and the arm was authored anyway: 2{,}787 word
types and 1{,}413 distinct openers, second to Kinyarwanda on both, from a brief
that offered nothing to copy.

\paragraph{A corpus authored cue by cue duplicates wherever its cue list does.}
Running the deduplication gate over the four arms separates them sharply and not
by language. Kinyarwanda carries \textbf{one} exact duplicate in 2{,}282
sentences (0.1\%). English, French and Kiswahili carry \textbf{98, 99 and 100}
in 2{,}400 (8.2--8.3\%). Every collision is a pair, never a larger group, and 97
of the pairs are \emph{the same two ids} in all three arms, which already rules
out an explanation in any one language.

The cause is upstream of all three. The three arms were authored against a
shared list of clinical cues, and that list contains \textbf{exactly 100
duplicate cues}: 2{,}300 distinct cues over 2{,}400 rows, no cue used more than
twice. Asked to write the same cue twice, the author writes the same sentence
about 96\% of the time (96 of 100 in English, 97 in French, 96 in Kiswahili),
and the duplicate rate of the corpus is therefore the duplicate rate of its
cue list, carried through almost intact. The remaining collisions come from cues
that are distinct but say the same thing, such as \emph{feels the coil string is
gone} beside \emph{cannot feel the coil thread any more}.

Kinyarwanda was authored phrase by phrase from a 165-phrase inventory rather
than cue by cue, and duplicates once. \textbf{The difference is the unit of
authoring, not the language, the author or the effort.} We report it because it
is a property of a construction method rather than of this corpus: a cue list
with repeated or paraphrased entries produces a corpus with repeated entries,
the repetition is invisible while the cues are read one at a time, and the only
place it surfaces is a gate that compares finished sentences.

One case is worth separating because it is a property of the target language
instead. Four Kiswahili collisions come from cues that differ in a feature the
sentence does not carry: \emph{woman + pain with a full bladder} and
\emph{old man + a pain that comes with a full bladder} are different cues, and
the Kiswahili sentence for both is \emph{Napata maumivu kibofu kikijaa}, because
it is first person and marks neither gender nor age. English and French collide
this way twice each. A cue can encode a distinction that a faithful sentence in
the target language has no way to express, and the corpus then looks redundant
where the authoring was not.

\paragraph{Two arms lifting the same eleven held rows is evidence about
concepts, not about a language.} The English and French briefs were built
against the same concept spine, each reviewed without reference to the other.
Of the \CorpusHeld{} held rows on the spine, English lifted 11 and French 12;
the two sets intersect in \textbf{11 rows across 7 concepts}, and French lifted
one row English did not (\texttt{EX27} third). The unit matters: eleven is a
count of held \emph{rows}, each a concept crossed with a grammatical person, not
of concepts.

For 6 of those 7 concepts the reasoning recorded on each side is the same and
was reached separately: the hold recorded a block on a Kinyarwanda \emph{word}
rather than on the concept the word expressed, so an arm not using that word had
nothing to be blocked by. \textbf{The seventh, \texttt{OB12}, is not independent
agreement and we do not present it as such}: the English arm ruled it and the
French brief records that it mirrored that ruling because it is a ruling about
the concept.

Both briefs also carry one hold the spine does not, on whether a patient who can
accurately report new confusion is meaningfully confused (\texttt{NE06} first).
\textbf{That is shared state rather than convergence.} The French record says in
terms that it carries the English arm's ruling, so it is one arm's clinical
question propagated, and it is not evidence of two arms agreeing.

Taking only the 6 concepts both arms reached separately, this is the one
cross-lingual result the project has, and it is a result about the taxonomy
rather than about any arm. A hold that survives translation into two unrelated
lexicons is a clinical question; a hold that dissolves in both is a lexical
one. Separating those two by hand is the kind of judgement a single
annotator cannot make about their own language, and here it fell out of building
the arms separately and comparing afterwards. We record the mechanism because it
generalises: where a taxonomy is shared but the wording is not, independent arms
are a cheap test of whether a recorded doubt belongs to the concept or to the
words.

\paragraph{The Swahili brief ships with no Swahili in it, deliberately.} The
sheet a Kiswahili speaker receives carries the English gloss, the person split,
the applicability and relation rulings and the open clinical questions, and an
empty column where the phrase goes. No machine draft is supplied, and a
machine-translated Swahili corpus that exists in the frozen v1 vocabulary is
withheld until the authored phrases exist.

This follows directly from what happened to the Kinyarwanda arm, which had to be
re-authored from scratch rather than corrected: a fluent draft anchors a reviewer
to its errors. Reviewing a plausible sentence is a different task from writing
one, and it produces a different artefact, because the reviewer's attention goes
to what is wrong with the draft rather than to what the patient would actually
say. We state it as a methodology position for any multilingual corpus effort:
if the arm is meant to be authored, supply no draft, and accept the slower start
as the price of not measuring a machine's register through a human's approval.
The cost is real and is visible in Table~\ref{tab:arms} as a zero.

\paragraph{Provenance of the phrases is recorded per phrase, and is mixed.}
Roughly half the 165 phrases are recorded as authored directly by the speaker;
the remainder are speaker-derived, machine drafts the speaker approved, or
machine-derived, with the disposition recorded per phrase in the repository.
\textbf{We state no percentage.} The per-phrase column exists, but no committed
script aggregates it, so any split quoted here would be \textbf{NOT
REPRODUCIBLE}; producing that aggregation is outstanding work.

This matters more than a footnote. The project's own working rules require that
speaker-authored and machine-drafted material stay distinguishable in the record
precisely so that a paper can report them separately, and elsewhere in this
repository the corpus is described as ``165 speaker-authored phrases''. Against
the per-phrase provenance column that description overstates the speaker's
contribution by roughly half, and we do not use it.
\paragraph{Flag counts, reconciled once.}
\label{sec:method:flagcounts}
Three flag counts appear in this paper and they are in different units, so the
reconciliation is given here and referred to rather than repeated.

\begin{itemize}\setlength{\itemsep}{1pt}
\item \CorpusFlagged{} \emph{rows of the authoring record} carry a flag for
  clinician review. This is the count Appendix~\ref{app:person} reports, over
  all \CorpusRows{} rows.
\item \CorpusFlaggedOnly{} of those carry a flag \emph{without} also being held.
  The other \CorpusFlaggedAndHeld{} are both flagged and held.
\item \CorpusFlaggedInInventory{} of the \CorpusInventory{} \emph{phrases the
  generator holds} carry a flag. This is smaller than \CorpusFlaggedOnly{}
  because two flagged, unheld rows still generate nothing: one is ruled
  inapplicable and one has not been written yet.
\end{itemize}

No phrase in the inventory carries a hold, because a held row generates nothing
by definition. The three figures are emitted from the record by
\path{review/emit_corpus_counts.py}, so they move together or the build fails.

\subsection{Concept taxonomy and clinical anchoring}

Phrases are organised under concepts, and concepts are anchored to published
clinical guidance wherever guidance exists for the presentation.

\paragraph{The concept total is not established, and we do not assert one.}\label{sec:method:concepts} Our
own audit found the concept count recorded with \textbf{five different values}
across this repository (68, 80, 126, 127 and 128) in documents that each
describe the same spine. A paper cannot quote one of five and call it a
measurement, so we quote none. Establishing the total against the authoring
record is outstanding work, and until it is done every count below is stated as
a count of \emph{anchor records}, which is what the anchor table actually
enumerates.

\paragraph{Anchor assignments are unverified, and we report no counts.} Concepts
were assigned anchors to published clinical guidance during corpus construction.
\textbf{Those assignments have never been checked against the source documents,
which this project does not hold}, so we give no count of anchored concepts, no
breakdown by document, and no per-concept reference. Earlier drafts stated a
count that its own table contradicted: the total silently absorbed a row of
concepts recorded as having \emph{no} anchor, which is the error this decision
removes. Verifying every assignment against the document and page it names is
outstanding work, and until it is done the anchoring is a property of our
construction process rather than a claim about clinical guidance.

The remaining spine positions are inherited from the v1 corpus, which
was catalogued without recorded concepts. They have no clinical reference and no
English gloss of their own; where this work needed a gloss for them it took the
frozen v1 English string at the same corpus position, which describes a
presentation rather than prescribing an utterance.

\paragraph{No clinician has reviewed the taxonomy.} The urgency assignments are
the author's reading of the guidance cited above, and nobody with clinical
training has checked them. We approached clinical contacts and did not secure
one; this is a gap we could not close, not a scope we chose. \CorpusFlagged{} rows carry an explicit flag for clinical review and
\CorpusHeld{} are held pending a decision; the review pack listing them, the five
open taxonomy questions and a sign-off sheet is written and unexecuted.

\paragraph{No second annotator has labelled any part of the corpus.} Every
urgency label in it is one person's. No second annotator has labelled any
part of it, so no inter-rater agreement statistic exists: not a low one, none.
The annotation protocol, the boundary cases we expect disagreement to
concentrate on, and the $\kappa$ computation are written and wait on a person.
We flag one design point in advance: agreement must be computed at the
\emph{concept} level, since rows are frame permutations and scoring them as
independent items would inflate any figure obtained.

No text is reproduced from any guidance document consulted during construction:
a clinical fact may be cited, the wording a document expresses it in may not be
copied. We name no document and quote no licence here, for the reason given in
Section~\ref{sec:related}: the assignments that would justify naming them have
never been checked against the sources. The licence analysis that produced the
no-reproduction rule is recorded in the repository.

\paragraph{Person, and the rulings that remove rows}

A speaker's rulings on grammatical person remove whole concepts and restrict which
relations may appear. The rulings, the concept-collapse table and the relation
restrictions are in Appendix~\ref{app:person}.

\paragraph{The other three language arms, as built}

The corpus reported here is Kinyarwanda; the frame fragments that wrap a phrase
are authored for that language alone, so \textbf{no other arm can generate a
single row}, and the component-by-component position is in
Appendix~\ref{app:arms}.

\subsection{Model, training configuration and environment}
\label{sec:method:training}

Stated in full so the run can be repeated. Every value is the one recorded in
the run's own report, not a value retyped from a configuration file.

\paragraph{Encoder.} \texttt{Davlan/afro-xlmr-mini}
\cite{alabi2022afroxlmr}, fine-tuned for
three-class sequence classification. The whole embedding module is frozen:
96{,}199{,}296 parameters of 117{,}641{,}859, being the
250{,}002\,$\times$\,384 word-embedding table (96{,}000{,}768) together with
the position and token-type tables and the embedding LayerNorm. The lowest 8
transformer layers are frozen as well, leaving 7{,}246{,}851 trainable
parameters in the reported configuration.

\paragraph{Hyperparameters.} Learning rate $1\times10^{-5}$; batch size 16;
maximum sequence length 96 tokens; one epoch with a 2{,}000-step cap, stopping
at 1{,}150 steps; warmup ratio 0.06; weight decay 0.01; seed 42 for both
training and the split. The reported checkpoint is step 900, selected on
stopping-set loss (0.5786), not the final step.

\paragraph{Environment.} Python 3.11; \texttt{torch} 2.12.0+cpu,
\texttt{transformers} 5.8.1, \texttt{scikit-learn} 1.8.0, \texttt{numpy}
2.4.4, all pinned exactly. CPU only; no GPU was used at any point.

\paragraph{Sequence length is a choice with a cost.} Kinyarwanda is
agglutinative and tokenises long. At 96 tokens some inputs truncate, and the
truncation rate per language is \textbf{NOT YET MEASURED}. A model that never
sees the end of a sentence cannot be said to have read it, so this is a
measurement we owe rather than a parameter we defend.

\subsection{Why the evaluation set must be the size it is}
\label{sec:method:power}

The figures 710, 365 and 720 come from exact binomial power calculations, and
the assumptions behind them are stated here because a sample size quoted
without its assumptions is exactly the kind of number this paper is about.

\paragraph{The common assumptions.} Confidence 0.95, two-sided, so 0.025 in
each tail. A threshold counts as met only when the \emph{lower} end of the
interval clears it, which is why an observed value exactly at the threshold
never passes. Power $\geq 0.80$, and required to hold at that $n$ and at the
next nine, so a minimum is not an isolated lucky point on the power curve.
Intervals are exact binomial (Clopper--Pearson), not normal approximations,
because the rates of interest sit close to 0 and 1 where the approximation
fails.

\paragraph{The three figures.}
\begin{itemize}
\item \textbf{710}, overall accuracy at a 0.82 threshold, assuming a true
  accuracy of 0.86.
\item \textbf{365}, CRITICAL recall at a 0.91 threshold in \emph{each} pure
  language, assuming a true recall of 0.95. The assumption is load-bearing: at a
  true 0.93 the requirement is 1{,}535, and at 0.97 it is 145.
\item \textbf{720}, the CRITICAL-to-ROUTINE rate below 0.01, assuming a true
  rate of 0.002. With zero observed events 368 would suffice; at a true 0.5\%
  it is 2{,}470.
\end{itemize}

\paragraph{Clustering is not modelled, and that makes these numbers
optimistic.} The calculations above treat items as independent. Our rows are
not: they are generated from a small inventory of seed phrases, so rows sharing
a seed are correlated, and the effective sample size is closer to the number of
distinct seeds than to the number of rows. No intra-cluster correlation
coefficient has been estimated, so no design effect is applied. \textbf{Every
minimum above should therefore be read as a floor on a floor.} This is the same
failure the corpus gates exist to prevent, appearing in the evaluation
specification instead of in the corpus.

\subsection{Split and leakage control}

The evaluation split holds out whole phrase groups, never rows. A phrase group is
the closure of a set of phrases under nesting: where one phrase is contained in
another, or is an ordered subsequence of it, the two are inseparable, because a
model trained on the longer string has seen every character of the shorter one
while an exact-match overlap check reports zero leakage.

Three silent attribution failures were found and fixed while building this, each
of which would have dropped rows out of the leakage analysis without raising an
error: a case-sensitivity mismatch, a placeholder-deletion bug that welded a
sentence's two halves together with a double space, and, the largest, a
phrase authored as a complete sentence never matching its own rendering, because
the generator removes the terminal stop before appending an onset. That last
reached most of the corpus, 57 of the 100 authored phrases at the time ending in
sentence punctuation, and it mis-attributed rows to a shorter phrase that happened
to be a prefix.


%

\section{Results}
\label{sec:results}

Each subsection states a finding, the measurement that establishes it, and the
script and report it comes from. All of it re-derives from a clean clone with
\texttt{make reproduce} (eight steps, verified on a machine with no prior state;
\path{ml_model/reproduce.py}).

\subsection{A corpus requirement stated in rows does not constrain quality}
\label{sec:res:seeds}

Our functional requirement asks for at least 1{,}000{,}000 unique examples.
Generating them took 130 seconds and passed the generator's own quality targets
(\path{dataset/generate_large_dataset.py}).

We then ran that corpus against the nine corpus-quality gates written in the same
specification (\path{dataset/corpus_gates.py}, 19 tests;
\path{reports/PRELIMINARY_RESULTS.md}). No gate was waived, and a gate that
cannot be computed from the data at hand is reported as such rather than as a
pass. Table~\ref{tab:gates} gives the outcome.

\begin{table*}[t]
\centering
\small
\setlength{\tabcolsep}{4pt}
\begin{tabular}{l>{\raggedright\arraybackslash}p{5.2cm}>{\raggedright\arraybackslash}p{9cm}}
\toprule
Gate & What it constrains & Outcome on our corpus \\
\midrule
C1 & rows per seed ($\leq 50$; no seed above 0.1\%) & \vFail: worst seed 8{,}975 rows, 26.07\% of the 34{,}425-row test split \\
\textbf{C2} & \textbf{distinct seeds per language ($\geq 3{,}000$)} & \vFail: 15 in the test split, 150 in train, 165 in the whole inventory. \textbf{This is the binding gate.} \\
C3 & lexical diversity against the pilot's floor & \vNC: the pilot that defines the floor does not exist \\
C4 & near-duplicate seeds ($\leq 2\%$ at Jaccard 0.85) & \vPass: 0 of 15 \\
C5 & machine person-transformation banned & \vFail: no recorded origin per row \\
C6 & frame consistent with reporter and age group & \vNC: neither is recorded \\
C7 & per-row provenance complete & \vFail: no seed, origin, author or validator fields \\
C8 & author concentration ($\leq 20\%$ per cell) & \vFail: the corpus has no authors \\
C9 & surface variation (case, punctuation, spacing, typos) & \vFail: 0 capitalised rows, 0 with stray spacing, no typos \\
\bottomrule
\end{tabular}
\caption{The nine corpus gates against the generated corpus. Four fail as
generated; six fail on the split whose rows can be attributed to a source
sentence, which is where C1, C2 and C4 become computable.}
\label{tab:gates}
\end{table*}

The binding gate, C2, counts \emph{distinct authored seed phrases} rather than
rows. Our generator has 165. Table~\ref{tab:seeds} follows that constraint
through.

\begin{table*}[t]
\centering
\small
\setlength{\tabcolsep}{4pt}
\begin{tabular}{r r >{\raggedright\arraybackslash}p{5cm} r}
\toprule
Seeds available & Rows allowed by C1 & Passes C2? & Largest passing corpus \\
\midrule
\textbf{165 (ours)} & \textbf{8{,}250} & \vFail\ (needs 3{,}000) & \textbf{0 rows} \\
3{,}000 & 150{,}000 & \vPass & 150{,}000 \\
20{,}000 & 1{,}000{,}000 & \vPass & 1{,}000{,}000 (the target) \\
\bottomrule
\end{tabular}
\caption{At 165 seeds no corpus of any size passes, because the floor is on
seeds and not on rows.}
\label{tab:seeds}
\end{table*}

%
%
\begin{figure}[t]
\centering
\begin{tikzpicture}
\begin{axis}[
  width=\linewidth,
  height=0.62\linewidth,
  xlabel={Distinct authored seed phrases},
  ylabel={Largest gate-passing corpus (rows)},
  xmin=0, xmax=20000,
  ymin=0, ymax=1050000,
  scaled y ticks=false,
  yticklabel style={/pgf/number format/fixed, /pgf/number format/1000 sep={,}},
  xticklabel style={/pgf/number format/fixed, /pgf/number format/1000 sep={,}},
  tick label style={font=\scriptsize},
  label style={font=\small},
  grid=major,
  grid style={gray!25},
  clip=false,
]
\addplot[thick, black, no marks] table[x=seeds, y=rows] {generated/seed_curve.dat};

\addplot[dashed, gray] coordinates {(\SeedFloorRaw,0) (\SeedFloorRaw,1050000)};
\node[anchor=west, font=\scriptsize, gray!60!black]
  at (axis cs:\SeedFloorRaw,980000) {\ seed floor \SeedFloor};

\addplot[only marks, mark=*, mark size=2pt, black]
  coordinates {(\SeedOursRaw,0)};
\node[anchor=south west, font=\scriptsize, align=left]
  at (axis cs:\SeedOursRaw,0)
  {ours: \SeedOurs\ seeds,\\[-2pt] 0 rows};
\end{axis}
\end{tikzpicture}
\caption[The seed floor and the per-seed cap, plotted.]{Table~\ref{tab:seeds} plotted. \textbf{This is arithmetic on the gate
thresholds, not a measurement}: the seed floor (\SeedFloor{} distinct authored
phrases per language) and the per-seed cap (\SeedRowsPerSeed{} rows) are written
in the corpus specification, and the curve is
$y = x \times \SeedRowsPerSeed$ where $x \geq \SeedFloor$ and $0$ below it.
Nothing here is fitted or observed. The discontinuity is the point: below the
floor the largest corpus that passes every gate is zero rows \emph{at any row
count}, so our \SeedOurs{} seeds sit on the flat section and no amount of
generation moves them off it. Data emitted to
\path{paper/generated/seed_curve.dat} by \path{review/emit_seed_curve.py}.}
\label{fig:seedcurve}
\end{figure}
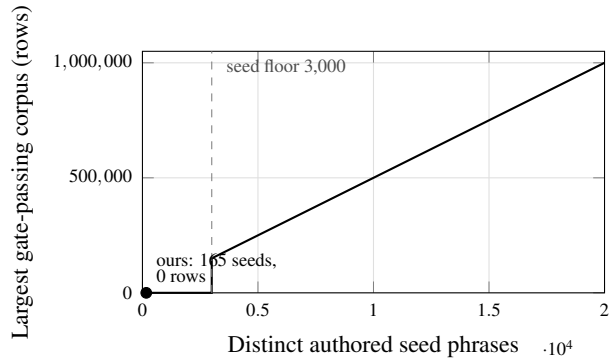

\paragraph{Two shortfalls, in sentences rather than rows.} A row is a frame
rendered around an existing phrase, produced at roughly 7{,}700 per second. A
seed is a sentence a native-speaking clinician writes. The two are often
conflated, including in an earlier draft of this paper, so we separate them.
\textbf{To pass the seed-count gate} the corpus needs 3{,}000 distinct seeds
against the 165 it has: a shortfall of \textbf{2{,}835}. \textbf{To reach the
one million rows the requirement names} it needs 20{,}000 seeds, because a
separate gate caps any one seed at 50 rows: a shortfall of \textbf{19{,}835}.
The gates do not demand twenty thousand sentences; the row target does. Reaching
that target requires 19{,}835 further seed phrases; at the authoring rates assumed in
our clinician brief (two to three minutes per item, an assumption and not a
measurement) that is on the order of 660 to 990 clinician-hours. Even at
20{,}000 seeds the generated corpus would still fail C5, C7, C8 and C9, which
constrain how a row is made (its origin, its provenance, its author, its
surface form) rather than how many rows exist.

\paragraph{The general claim.} Expansion is the cheap operation in any templated,
augmented or model-expanded corpus, and a requirement written in rows constrains
exactly that. The quantity that resists expansion, and that the model needs, is
the number of independently authored source items. A specification that sets no
floor on that number has not constrained quality, however large the row count it
demands. The remedy is small: constrain seeds, and let the row count follow.

\takeaway{A row count constrains the cheap quantity.}{Rows are
produced by machine at roughly 7{,}700 per second; seeds are written by
native-speaking clinicians. A requirement stated in rows therefore binds the
free quantity and leaves the expensive one unconstrained, which is why a corpus
can meet it and fail four of the nine gates written beside it.}

\subsection{Cluster structure, not row count, decides what a held-out set can show}
\label{sec:res:eval}

Our evaluation set is 17{,}942 rows built from nine distinct source sentences,
four of them carrying the critical label (\path{reports/DATASET_AUDIT.md}).
Our deployment gate counts distinct source sentences rather than rows and
computes its intervals by resampling sentences
(\texttt{training/evaluate.py}). On this set it refuses 38 of its cells and
reports no metric at all; the refusal is reproduced from a clean clone as step
seven of \texttt{make reproduce}.

The required sizes are derived rather than asserted
(\path{training/eval_spec.py}, whose power calculations are recomputed by a
test): 710 distinct sentences for overall accuracy, 365 gold critical items per
pure language for critical recall at the stated threshold, and 720 for the
critical-to-routine rate. Nine sentences is not a small sample of the required
set; it is three orders of magnitude below it on the axis that matters.

\paragraph{Two sets, and which is which.} Three names appear in this paper for
two sets, and they are not interchangeable.

\begin{itemize}
\item The \textbf{test split}: 34{,}425 rows built from 15 distinct seed
  phrases. This is what the corpus gates in Table~\ref{tab:gates} are measured
  against.
\item The \textbf{reporting set}: 17{,}942 rows from 9 distinct sentences, in 5
  phrase groups. Every per-class metric in this paper is measured on this.
\end{itemize}

The second is a subset of the first. Three phrase groups of the test split are
held out as the stopping set during training, so the reporting set is the test
split minus the sentences whose loss selected the checkpoint. Reporting on the
whole test split would mean reporting on rows that chose the model.

That is why a share of one is not a share of the other: the largest seed is
26.07\% of the test split and 50.02\% of the reporting set, and quoting either
figure without naming its denominator makes the concentration look like a
different problem than it is.

\paragraph{Seed provenance, stated before any metric.} Train and test share no
seed: 150 distinct source phrases and 295{,}575 rows against 15 and 34{,}425,
with zero shared seeds and zero test rows built from a shared seed
(\path{dataset/seed_provenance.py}). Disjoint is not independent. Both splits
come from one generator and one 165-phrase inventory, so any metric measured
across them would describe generalisation within the generator's distribution.
The largest single seed accounts for 26.07\% of that 34{,}425-row test split.

\paragraph{The floor a number would have to clear.} A constant classifier that
always answers the most frequent class scores 0.4995 on that evaluation set and
0.3418 on the wider generated split (\texttt{reports/measurements/}
\path{majority_baseline.py}, labels only, no model). We record these so that
any future number is read against them.

\paragraph{A single-condition gate certifies the wrong model.} Our first
acceptance gate was one condition, recall on the critical class, and a
configuration passed it while abandoning the urgent class entirely, scoring on
the protected class below what merging the two classes earns for free on the same
rows. The figures, their closed form and the three-condition gate that replaced
them are in Section~\ref{sec:disc:gate}; they are a defect in gate design and
support no quality conclusion about any configuration. All three configurations,
the per-class figures and the confusion matrix are in
Appendix~\ref{app:run}, so the claim can be checked without re-running the
pipeline.

\takeaway{Count clusters, not rows.}{17{,}942 rows built from nine
distinct sentences are nine observations wearing a large number. Every interval
is governed by the distinct-sentence count, so a gate that counts rows will
report confidence it has not earned. Ours counts sentences and refuses.}

\subsection{Two failure modes that leave every output-side signal healthy}
\label{sec:res:silent}

\paragraph{A model shipped without its tokenizer.} Loading a model directory that
contains no tokenizer files does not raise. The loader returns a vocabulary of
five tokens, every word becomes the unknown token, and the model answers its
class prior on input it cannot read. Probabilities remain well formed, predictions
remain deterministic, and the label order remains correct. We encountered this in
our own probing harness and read it as single-class collapse until it contradicted
a previously recorded confusion matrix. No metric computed from the model's own
outputs distinguishes the two cases. Our probe now measures the vocabulary size
and the unknown-token rate of the tokenizer it actually loaded and fails loudly
(\texttt{training/probe.py}; \path{reports/MODEL_AUDIT.md} §11.1).

\paragraph{Surface fragility.} The same probe measures whether the predicted
class survives surface variation. On 200 sampled inputs, \textbf{drawn from 9
distinct source sentences with no interval reported, because 9 clusters cannot
support one}, the predicted urgency changes for 31.5\% under capitalisation and
21.0\% under a single realistic typo, for 5.0\% when punctuation is removed, and
for none under extra whitespace.

\textbf{We do not claim the corpus caused this.} The corpus does contain zero
capitalised rows, zero rows with stray spacing and no typographical noise at all
(Table~\ref{tab:gates}, C9), and that is a plausible contributor. But at least
two other explanations are available on the same evidence, and nothing we
measured separates them:

\begin{itemize}
\item the tokenizer is \textbf{cased}, so capitalisation changes the token
  sequence before the model sees it. That is a different input, not a fragile
  response to the same one;
\item the artefact's confidence \textbf{never exceeded 0.55} on the critical
  class (Section~\ref{sec:future:calibration}), so it sits close to its own
  decision boundary throughout, where small input changes flip the argmax for
  reasons that have nothing to do with the training data.
\end{itemize}

Whitespace flipping for none is consistent with the tokenizer normalising it
away, which is a fact about the tokenizer rather than evidence about the corpus.
Separating these would need a probe against a cased and an uncased tokenizer, or
a model whose confidence is not pinned near the boundary; we ran neither.

We therefore report the flip rates as a measurement and not as an attribution. A
patient typing in capitals or mistyping a word is not making a clinical
statement, so we record this as a corpus-health signal and not as a quality
metric, and we do not propose normalising input at serving time: that would hide
the fragility and discard information the tokenizer is case-sensitive to.

\takeaway{A healthy output signal is not evidence of a healthy
model.}{A missing tokenizer and a fragile decision boundary both leave
probabilities well formed, predictions deterministic and label order correct.
Each was caught only by checking a property outside the model: the encoding of
its input, and the invariance of its answer under a change that should not
matter.}

\subsection{What we do not claim}
\label{sec:res:none}

\textbf{No figure in this paper is offered as evidence of model quality.} That
is a narrower statement than the one an earlier draft made, and the narrower one
is the true one: model numbers do appear here, because several of the failures we
report are failures \emph{about} those numbers, and a paper cannot describe a
gate certifying a degenerate model without saying what the gate saw. A recall of
0.9749 beside an urgent recall of 0.0083 is evidence about a gate, not a
performance claim, and we make none.

No evaluation set meeting our specification exists; our gate refuses on the set
that does; and no model has been trained through the pipeline that would produce
a calibrated, threshold-tuned checkpoint. When run, it refuses before reading
the training corpus, for the reasons in §\ref{sec:res:eval}. The trained artefact
that exists is an audit artefact used to exercise the serving path and the gate's
refusal. It is not a baseline: we do not compare against it, and we do not plan to
retrain it.

%
%

\section{The apparatus}
\label{sec:system}

This section describes what the system does today. None of it is evidence that
the system works: no model has passed our gate, and the properties below are the
ones we built so that a failing model fails safely rather than quietly. We state
them because a reader assessing the negative results needs to know what the
apparatus does when the model is wrong, absent, or unfit, which is its normal
condition at present.

\subsection{Safety properties of the apparatus}

\paragraph{It fails closed.} When no model is loaded, or the model cannot
classify, or inference times out, the triage endpoint returns an error carrying a
manual-triage instruction to staff, and \emph{nothing is written}: no symptom
report, no urgency, no queue entry. An earlier build served a keyword matcher
whenever no model was configured. Measured on the frozen Kinyarwanda holdout it
caught a small minority of critical cases and sent the large majority of them to
the routine queue with a confident message; the exact rates came from a Phase~0
script that was never committed and are therefore \textbf{NOT REPRODUCIBLE}, so
we state the direction and not the figures. A classifier that is wrong in the dangerous direction is
worse than no classifier, because nobody triages a patient the system has already
triaged. The fallback was removed, and the absence of a model now produces a
refusal rather than a guess.

\paragraph{It never tells a patient their condition can wait.} Patient-facing
text is identical in structure across all three urgency classes and ends with an
escalation instruction in every case. No urgency class produces a reassuring
message, because the label is a queue-ordering hint and not a clinical judgement,
and a patient reading ``routine'' as ``safe to stay home'' is a harm the system
would have caused. Patients receive a receipt, a queue position and that
instruction, and nothing else.

\paragraph{Cases the model could not classify sort above urgent and routine.}
The queue is ordered critical first, then cases the model declined or answered
below the review threshold, then urgent, then routine. An unclassifiable case is
not a low-priority case: it is a case about which the system knows nothing, and
placing it behind cases the system does have an opinion about would use absence
of evidence as evidence of safety.

\paragraph{A deterministic rules layer runs before the model, and can only
escalate.} Terms in the layer force the critical label regardless of model
output; the database rejects any row in which the layer lowered an urgency, so
the escalate-only property is enforced by a constraint and not by convention.
\textbf{The term table ships empty}, because no clinician has validated a term
list, and an unvalidated rule creating critical queue entries would be a new
unvalidated claim rather than a safety net. The layer therefore changes nothing
today. We describe it because the empty table is the honest state of it, and
because the enforcement is what makes filling it safe later.

\paragraph{Configuration that would silently change the served decision is
refused at start-up.} The service reads the length the model was fine-tuned at,
the calibration temperature and the decision thresholds from the model directory,
and applies them exactly; a record it cannot apply exactly (half a rule, an
unknown rule text, a different label order, a configured length that disagrees
with the recorded one) stops start-up rather than becoming a quiet error at a
patient's request. This exists because the decision a model is tuned for, the
decision a gate scores, and the decision a service serves are three separate
things that can drift apart silently; we found two such gaps in our own system
and closed both.

\subsection{An invariance check the fragility finding implies}
\label{sec:system:invariance}

\textbf{Proposed, not implemented, and not evaluated.} No code in this
repository does what this subsection describes, no measurement of its effect
exists, and nothing below should be read as a property of the served system. We
state it because the surface-fragility result in
Section~\ref{sec:res:silent} implies a countermeasure, and a reader who accepts
that result will reasonably ask what follows from it.

The measurement was that the predicted urgency changes for 31.5\% of inputs
under capitalisation and 21.0\% under a single typo. Whatever the cause, and
Section~\ref{sec:res:silent} declines to attribute one, a prediction that moves
under a change the patient did not intend is not a prediction the serving path
should act on silently.

\paragraph{The check.} Before serving a triage decision, the endpoint copies the
input, applies the surface perturbations the probe already implements (case
folding and punctuation removal), runs the model on the original and the
perturbed copy, and compares the argmax. If the two agree, it serves the
decision. \textbf{If they differ, it refuses rather than serving}: the request
is routed to human review with the disagreement recorded, exactly as the gate
refuses a cell it cannot measure. The system already fails closed elsewhere, so
this is the same posture applied one layer further out.

\paragraph{Its cost.} Two forward passes per request instead of one, so
throughput halves and latency roughly doubles. On a task whose latency target is
already recorded in this paper as unachievable on the hardware we measured, that
cost is not obviously affordable, and we do not claim it is. It would need to be
measured against a real target machine before anyone adopted it.

\paragraph{What it does not do.} \textbf{It catches instability, not
wrongness.} A model that is stable and confidently wrong passes it unchanged:
both passes agree, the check is satisfied, and the wrong urgency is served with
no flag. It is a consistency test, not a correctness test, and it would be a
mistake to read a low disagreement rate as evidence that the model is right. It
also cannot detect a failure that affects both copies equally, which includes the
missing-tokenizer mode reported beside it: a model answering its class prior
answers the same prior twice.

\subsection{What the apparatus has not been shown to do}

It has not run in a health centre and has not been used with any patient. No
community health worker or clinician has been consulted about how patients
present. The latency of the served model was measured, but on a laptop that is
not a named target machine, against a target that our own audit records as
specified with no stated source and no check that it was achievable on any named
hardware. We therefore \textbf{report no latency result} for the served model,
and treat the target itself as one of the unchecked numbers this paper is about.

What we can say is about the \emph{target}, not about our model: on the one
machine we measured, meeting the specified figure at the stated concurrency
would require roughly 26 times the throughput that machine provides. That is a
statement about an unachievable requirement, not a performance measurement, and
it is not offered as one.

\paragraph{Code-switching: none of the six combinations generates}

The corpus design names six ordered language pairs and \textbf{none of them
produces a row today}. A speaker ruled fifteen medical terms for insertion, which
yielded four cleanly switched terms and five borrowings that belong in monolingual
rows instead, too thin an inventory to be the mixed language the label claims. The
generator therefore declines a pair it cannot integrate morphologically rather
than approximating one, and the full position, pair by pair, is in
Appendix~\ref{app:codeswitch}.

%
%

\section{Discussion}
\label{sec:discussion}

\subsection{Why a row count cannot constrain a corpus}

The finding in Section~\ref{sec:res:seeds} is easy to read as a local accident of
our own specification. It is not, and the reason is an asymmetry present in every
pipeline of this kind.

\paragraph{The two quantities have costs that differ by orders of magnitude.} A
row is produced by rendering a frame around an existing phrase: our generator
emits roughly 7{,}700 per second, and the marginal cost of the millionth row is
indistinguishable from the cost of the first. A seed is a sentence someone who
speaks the language writes, at minutes each, and no amount of compute
substitutes. A requirement that names the first quantity therefore constrains
nothing that is scarce, and one that names the second constrains the thing that
actually bounds what a model can learn.

\paragraph{The substitution is invisible in every aggregate a reviewer normally
sees.} Row count, class balance, language balance, mean length and duplicate rate
all look healthy on our corpus: the generator's own quality targets passed
before any gate ran. What fails is a property no aggregate exposes: how many
independent things the corpus says. A reader cannot see it without being told the
distinct-seed count, which is why we state it at the head of every report and
beside every figure.

\paragraph{The general form.} For any templated, augmented or model-expanded
corpus, expansion is cheap and authoring is not; so a requirement written in the
expanded unit is satisfiable without producing any of the scarce unit. The remedy
is not a larger row target. It is to write the requirement in the scarce unit,
distinct authored items per language and per cell, and let the row count follow
as a consequence. We would apply the same test to any corpus claim we read: ask
what the expensive quantity is, and whether the stated requirement mentions it.

\paragraph{Where the argument stops holding.} Seeds-not-rows holds where frames
\emph{permute} rather than \emph{vary}: where the expansion step recombines a
fixed inventory of wrappers around an invariant seed, so that expansion adds
tokens and adds no information. Two conditions break it. The first is a generator
whose frames introduce genuine linguistic variation, changing syntax, register or
propositional content rather than decorating it; there the expanded items are not
permutations of one observation and the effective sample size is larger than the
seed count, though how much larger is an empirical question and not one a row
count answers either. The second is a task where surface form itself carries
label information, such as register classification or authorship attribution:
there a frame change is a legitimate new observation, because the frame is part
of what is being predicted. Our corpus meets neither condition, and a reader
applying the argument elsewhere should check both before assuming it transfers.

\subsection{Why cluster structure, not row count, governs an evaluation set}

The same asymmetry decides what an evaluation set can establish, and here it has
an exact statistical form rather than an economic one.

\paragraph{Rows from one sentence are not independent observations.} Our
evaluation set holds 17{,}942 rows built from nine distinct sentences. Frame
permutations of one sentence share its vocabulary, its syntax and its label;
whatever the model has learned about that sentence, it applies to all of its
permutations at once. The effective sample size is therefore governed by the
number of clusters, not the number of rows, and the discrepancy here is three
orders of magnitude.

\paragraph{An interval computed over rows is not merely optimistic; it is
answering a different question.} Resampling rows estimates how the number would
move if we drew more permutations of these nine sentences, a quantity nobody
wants to know. Resampling sentences estimates how it would move on nine different
sentences, which is the question a deployment decision asks. Our gate therefore
counts distinct sentences per cell, resamples sentences for every interval, and
refuses a cell whose distinct-sentence count falls below the size its threshold
requires. On this set it refuses 38 cells and reports nothing, which is the
correct output rather than a failure of the instrument.

\paragraph{What this implies for low-resource evaluation generally.} A test set
assembled by expanding a small seed inventory can be made arbitrarily large and
will not become more informative. Where a corpus is built this way (which in
low-resource settings it very often is), the honest reporting unit is the
distinct source count per cell, and a paper that reports only row counts has not
told the reader what its numbers rest on. The cost of the guarantee is real and
should be stated openly: unioning a concept's phrasings into one cluster, as
Section~\ref{sec:disc:leakage} describes, reduces the number of independent
evaluation units, and that reduction is the price of a leakage guarantee rather
than an accident of it.

\subsection{Why a single-metric gate fails, and what replaced ours}
\label{sec:disc:gate}

The gate below is the project's own acceptance gate, generated from the record
that defines it. Two points about its numbers, since this paper is largely about
numbers without derivations. Its critical-recall threshold of 0.95 is
\emph{inherited and unverified}: a different figure from the 0.91 our
requirements document sets, and neither has a source we could trace. Its
non-degeneracy floor is derived exactly, though it guarantees less than it
appears to: macro F1 above $2/3$ implies only that the weakest class has
$F1 \geq 3\,\mathrm{macro} - 2$, which is zero at the threshold itself. Its
margin above the merge-strategy precision is a clinical judgement nobody has
made. The gate states which is which, which is
the property we would ask of any gate.

\begingroup
\renewcommand{\subsection}[1]{}
\renewcommand{\subsubsection}[1]{}
%
%
\subsection{The acceptance gate, and where each threshold comes from}
\label{sec:gate}

A model is accepted only if all three conditions hold. They are not traded off against one another.

\paragraph{A1: CRITICAL recall $\geq 0.95$.} \emph{Inherited; source not verified.} This threshold predates the current record and carries only the note that missing a critical case is the failure that matters. It is consistent with the trauma field-triage convention of holding under-triage at or below 5\%, but we have not verified that attribution against a source and do not claim it. We keep the value because loosening a safety threshold on no evidence is worse than retaining an unsourced one.

\paragraph{A2: macro F1 $> 2/3$, strictly.} \emph{Derived exactly.} Macro F1 is the unweighted mean of three per-class F1 scores. If any one class is abandoned its F1 is zero, so macro F1 $\leq (1+1+0)/3 = 2/3$ \emph{even when the other two classes are perfect}. The comparison is therefore strict: a model scoring \emph{exactly} $2/3$ is consistent with two perfect classes and one dead one, and an inclusive $\geq$ admitted it. \textbf{What the condition actually guarantees} is weaker than non-degeneracy in any useful sense: from $f_i \leq 1$ it follows that $\min_i F1_i \geq 3\,\mathrm{macro} - 2$, which is $0$ just above $2/3$ and $0.3172$ at the $0.7724$ measured here. Passing A2 proves no class is exactly dead; it does not prove no class is nearly dead. A model at macro $0.672$ may carry a class at $F1 = 0.016$ and pass.

\paragraph{A3: CRITICAL precision $\geq$ the merge-strategy precision plus a margin.} \emph{Floor derived and measured per evaluation set; margin not derived.} A model that merges CRITICAL and URGENT and labels the union CRITICAL earns, by construction, a CRITICAL precision of
\[ \resizebox{\columnwidth}{!}{$\displaystyle \frac{n_{\mathrm{CRIT}}}{n_{\mathrm{CRIT}} + n_{\mathrm{URG}}} = \frac{8{,}962}{16{,}755} = 0.5349$} \]
on this reporting set, where $n$ is a count of gold rows. \textbf{No model is involved in this quantity}: it is the CRITICAL class prior within the two urgent classes, computed from support counts alone, and it is the score the merge strategy is paid for free. \textbf{It is the precision of one specific strategy, not a ceiling on degenerate models in general}, and it assumes no ROUTINE row is labelled CRITICAL; a strategy that swept ROUTINE in as well would score lower. It is what that strategy is paid for free, so an informative model must exceed it. We compute it against the set being scored rather than hardcoding a value, because it moves with the set's composition. \textbf{The condition is not always satisfiable.} When CRITICAL rows dominate the two urgent classes the merge precision approaches $1$, the floor exceeds $1$, and no model can pass; the gate reports NOT COMPUTABLE rather than failing a model for a property of the set it was scored on. \textbf{The margin above that floor (0.10) is not derived.} How far above provably-degenerate a deployable model must sit is a question about tolerable over-triage in a clinic, it is a clinical judgement that has not yet been made, and we report it as a placeholder rather than as a standard.

%
%
\subsection{A single-metric safety gate certified a model that had abandoned an urgency class}
\label{sec:gate-failure}

Our acceptance gate was initially a single condition, CRITICAL recall $\geq 0.95$, on the reasoning that missing a critical presentation is the failure that matters. One configuration passed it with a CRITICAL recall of 0.9749 while achieving an URGENT recall of 0.0083: it assigned the CRITICAL label to 7,633 of 7,793 URGENT rows. It passed the safety gate \emph{because} it over-triaged, and the gate contained no term that could observe this.

The degeneracy is measurable rather than interpretive. The quantity it is measured against is a \textbf{class prior, not a model output}: the CRITICAL share of the two urgent classes' gold support, which any strategy that merges CRITICAL and URGENT and labels the union CRITICAL earns by construction, without learning anything. On this set that is a CRITICAL precision of $0.5349$ on this set by construction. The model scored $0.5337$, which is $0.0012$ \emph{below} its own set's merge-strategy precision. Within a thousandth, it did not approximate the merge strategy; it was the merge strategy.

Two further observations. First, the gate certified this model while rejecting a better one: a configuration with macro F1 0.7724 and all three classes alive failed the same gate on recall alone. Second, our own automated degeneracy check, written for exactly this failure, used the test $\mathrm{recall} \geq 0.9 \wedge \mathrm{precision} < 0.5$ and did not fire, because $0.5337$ sits above a hardcoded $0.5$ while the true merge-strategy precision for this set was $0.5349$. The lesson is not that the constant was too low: it is that the quantity is a property of the evaluation set and must be computed against it.

\endgroup

The specific failure is easy to dismiss as a badly chosen threshold. The general
form is not particular to triage.

\paragraph{A single-metric gate on an asymmetric-cost task is satisfiable by
collapsing the classes it protects.} Our gate was chosen for a defensible
reason: missing a critical presentation is the failure that matters, so hold
recall on that class. But recall on one class is maximised by predicting that
class, and a model that does so passes while destroying every distinction the
system exists to make. The gate did not fail to notice over-triage; it had no
term in which over-triage could be expressed.

\paragraph{The merge strategy's precision has a closed form and should be
computed rather than guessed.} What made this diagnosable rather than arguable is
that merging two classes and labelling the union with the protected class scores a
predictable precision on any evaluation set, namely the protected class's share of
the merged support. \textbf{That share is a class prior, computed from gold
support counts with no model involved}, which is what makes it a baseline rather
than a rival result: on our reporting set it is
$8{,}962 / 16{,}755 = 0.5349$, and the configuration we rejected scored
\emph{below} it. We could therefore say not that the model \emph{resembled}
the degenerate strategy but that it sat below that strategy's own precision on
those exact rows. This is the precision of one specific strategy and not a
ceiling on degenerate models in general: a strategy that swept the routine class
in as well would score lower. We would report the quantity beside any
single-class recall figure: it is defined for any confusion matrix and converts a
judgement call into arithmetic.

\paragraph{A hardcoded degeneracy guard drifts out of range.} We had written an
automated degeneracy check before any of this happened, using a fixed precision
threshold. It did not fire, because the merge strategy's precision on that
evaluation set was above the constant we had chosen. The lesson is not that the
constant was too low but that it was a constant: that precision moves with the
evaluation set's class composition.

\paragraph{The gate certified the worse of two configurations.} The rejected
configuration kept all three classes alive; the accepted one had abandoned an
urgency class entirely. We make no claim about which would perform better
clinically (both were scored on nine sentences), but a gate that prefers the
one that has stopped discriminating is not measuring what it was written to
protect. Any process that trusted the gate (an automated
promotion step, a continuous-integration check, a tired reviewer) would have
shipped the wrong model with a passing test to point at.

\paragraph{The leakage guarantee rests on a declaration, not on a similarity
rule}
\label{sec:disc:leakage}

Phrase-group closure computed from the strings is not sufficient, and no
similarity threshold could have made it so, so the guarantee rests on authoring
metadata instead; the mechanism, its \textbf{NOT REPRODUCIBLE} supporting figure
and the same shape recurring three times are in Appendix~\ref{app:leakage}.

\paragraph{The cost of the configuration search}

Three configurations were trained to completion at a total cost under five hours
on two CPU cores; the cost, what it does and does not establish, and why a budget
that small is the right size for 150 distinct training phrases are in
Appendix~\ref{app:search}.
\subsection{Recorded but unread}
\label{sec:disc:unread}

Thirteen times in this project something already known went unread until an
unrelated change forced a look. They fall into four kinds. Six are a fact
recorded correctly, in a machine-readable place, that no code path consumed.
Four are \emph{instruments} that reported on something they had not read. Two
are a count or a sentence that was correct when it was written, invalidated by a
change to the thing it described, and never re-derived. The last is a kind of its
own and the most uncomfortable: an instrument that read correctly, reported
correctly, and was understood to have answered a broader question than the one
it was asked. The list below is chronological, so the kinds are interleaved
rather than blocked. None is a measurement failure: in every case the
measurement was right and available, and the gap is between having a fact and
having something that reads it for what it says.

\begin{itemize}\setlength{\itemsep}{1pt}
\item \textbf{The split manifest.} It recorded that 100\% of one split's
  evaluation rows were built on phrases also present in training, in a field, in
  the commit that created the pipeline. The two fields anyone read were both
  honestly zero (Appendix~\ref{app:leakage}).
\item \textbf{The duplicated threshold.} One acceptance constant existed as
  independent copies in two source files, so changing the safety threshold in one
  would silently not apply in the other (Appendix~\ref{app:leakage}).
\item \textbf{The emitted tables nothing included.} Three tables were written
  by the evaluation emitter on every run and \texttt{\textbackslash input} by no
  file. The build header had said since the pipeline was written that the results
  section inputs them. Nothing failed, because a generated file nobody reads
  produces no error (Appendix~\ref{app:run}).
\item \textbf{The Swahili brief's own count.} A docstring said eleven lifted
  holds, the constant beside it held twelve, and the run's summary printed twelve.
  The validator checked that constant against one of the two prior arms, and a
  check against one side cannot detect a disagreement between two sides.
\item \textbf{A gate that could not fail.} One of the nine corpus gates
  returned a fixed \emph{not computable} string and read no column at all. It
  gave the same answer whatever it was given, including a sentence asserting that
  a corpus carried no reporter or age field about a corpus that carried both. It
  ran, printed a verdict, and was counted among nine.
\item \textbf{The gloss field.} The authoring sheet carries an English gloss per
  row, and the sheet requires it. In the reporting split this paper's figures rest
  on, 2 of 15 phrases carry a real gloss and the remaining 13 carry a placeholder
  telling the author where to look. Nobody read the field until an emitter needed
  one for Table~\ref{tab:example}.
\item \textbf{A checker silenced by the change it existed to verify.} We wrote a
  script to report the float type and column spec of every table, after a
  single-column float printed over the body text. Converting one table to
  \texttt{tabularx} to fix a width problem put it outside the one environment
  name the script matched, and it reported that table as zero columns and no
  content, which its own summary counted as passing.
\item \textbf{A gate whose floor does not exist.} C3 constrains lexical
  diversity against a threshold defined as a fraction of a natively authored
  pilot's diversity. No pilot has been run. The gate therefore cannot constrain
  anything, and reports NOT COMPUTABLE rather than saying that its own
  definition is unsatisfiable. It shares its shape with the stub above: both
  look like constraints in a list of nine and neither is one, and \textbf{neither
  would have been caught by running them}, because a gate correctly declining on
  thin data and a gate that can never do anything else produce the same word.
\item \textbf{A collapse detector that read the wrong key.} A sweep harness was
  given a function to report when a trained arm predicts a single class for every
  input, because a degenerate arm is a result and not an error. It read a
  top-level \texttt{confusion} field. The trainer writes the matrix at
  \texttt{metrics.confusion\_matrix}. The function would have returned
  \emph{no collapse} for every arm in the sweep, including a collapsed one,
  and nothing would have contradicted it.
\item \textbf{A file count in a commit message.} A rebuilt archive was described
  as holding 32 files. It held 29, plus two directory entries. The 32 was true of
  the archive that existed when it was written; excluding a 138\,KB plain-text
  rendering of the paper, which is not a source the archive needs, dropped the
  count, and the number was carried into the message without being re-derived
  from the archive it described. The archive was correct. Only the
  sentence about it was wrong, and it is a sentence no build step reads.
\item \textbf{Freshly emitted numbers in prose written for different ones.} The
  caption of Table~\ref{tab:example} reports how many phrases in the reporting
  split carry a real English gloss. A constant named \texttt{PLACEHOLDER} held
  the gloss placeholder; a second constant of the same name was added sixty lines
  below it for an unrelated purpose and silently shadowed the first, so the test
  for a real gloss became a test no gloss fails and the count went from 2 to 15.
  The emitter then substituted 15 into a sentence composed for 2, producing a
  caption that asserted in consecutive clauses that 15 of 15 phrases are glossed
  and that \emph{every other phrase} is not. The emitter did its job. Nothing
  checked that the sentence still parsed against its own values.
\item \textbf{A digest that was stable because the gap was.} The paper's
  submission archive was assembled by a shell command naming each file. One of
  them, the bibliography that the submission server will not regenerate, did not
  exist; the archiver said \emph{name not matched} into a pipe whose output was
  discarded, and built the archive without it. Two commits shipped that way. The
  archive's SHA-256 was recomputed on each rebuild and matched every time, and
  that stability was read as reproducibility. It was: the same absence was being
  reproduced exactly. A digest answers whether two things are identical and
  never whether either is complete, and we had been using it for the second
  question.
\item \textbf{A font check read one column wider than it ran.} Preparing to
  match the presentation of a paper we had modelled ours on, a reviewer ran
  \texttt{pdffonts} over both PDFs, saw Computer Modern named on each, and
  recorded that the fonts already agreed and should not be touched. Every word
  of that was true about the typeface. \texttt{pdffonts} also prints a
  \emph{type} column, and ours read Type~3: bitmapped fonts, which is what
  \texttt{fontenc} falls back to when no scalable font is loaded beside it. The
  same typeface can embed either way, so the column that was read could not
  distinguish the two cases and the column that could was not. This is the only
  instance here contributed by a reader rather than by the code, and it is the
  one we would generalise furthest: the tool was correct, the output was
  correct, and the defect was in taking an answer about one property for an
  answer about the thing.
\end{itemize}

\paragraph{Four of them are a different animal, and worse.} Six are records
that drifted from the thing they described: a field, a constant, a docstring or
a gloss that stopped matching reality while both sat still. Four are
\emph{instruments} that reported on something they had not read: one silenced by
the very change it existed to verify, one reading a field name that does not
exist, one whose threshold is a fraction of a study nobody has run, and one
whose digest confirmed that an incomplete archive was identical to itself. A table it could not parse scored identically to a table with nothing
wrong, because the script reported what it found and had no way to say
\emph{I did not understand this one}.

\textbf{A checker that passes what it cannot parse is worse than no checker},
and the reason is not subtle: absence of a checker is visible, and a green result
from a blind one is not. It answers the question it was asked with the word the
reader wanted. We caught the seventh only because the count of columns it printed
was obviously wrong for a table we had just written, and a less conspicuous
failure would have survived. The lesson we take is that a verification script
needs a third outcome beside pass and fail, and must treat \emph{unparsed} as a
failure rather than as silence. The collapse detector now raises when the matrix
is absent instead of returning \emph{no collapse} from a matrix it never read.

\textbf{The eighth is the one we would ask a reader to weigh, because of when it
happened.} It was written within an hour of the paragraph you are reading, in
code whose purpose was to catch a different failure, by someone who had just
finished naming the pattern. Knowing about a failure mode, having written it
down, and actively looking for it did not prevent committing it again. We report
that plainly rather than presenting the pattern as one we have now escaped: the
evidence of this section is that naming a recurring error does not inoculate
against it, and the only thing that has reliably caught these is a test that
fails when a value and its consumer disagree.

\paragraph{The thirteenth was created by a fix that was right.} It is worth
stating on its own, because it is the least comfortable thing in this section
and the easiest to leave out. The bitmapped fonts were not an oversight that
survived review; they were introduced \emph{by} a deliberate, correct repair.
T1 font encoding was added because \texttt{\_} was not reliably available as a
glyph and the compiled PDF was dropping it from every file path it printed: a
paper arguing that a reader must be able to re-run the code was naming the
scripts under names that do not exist. Loading T1 fixed that. It also asked for
a font nothing in the preamble supplied, and the fallback embeds as bitmaps.
No error, no warning, no failing check, and a correct fix strictly better than
what preceded it in the respect it was made for.

We draw a narrow conclusion rather than a broad one. It is not that fixes are
dangerous. It is that a change can be right about its own subject and wrong
about a property nobody was looking at, and that the properties nobody is
looking at are not discoverable from inside the change. Ours was found by
comparison with an artefact made by someone else.

\paragraph{Why it recurs, and what would actually help.} The common cause of the
first six is not carelessness about recording. It is that a project accumulates rulings faster
than it accumulates places to read them: every new decision adds a field, a
constant or a file, and each one needs a consumer that fails when the value is
wrong. Writing the fact down is cheap and feels like completion; wiring a reader
to it is the expensive half, and it is the half that gets deferred.

The pattern is visible only in hindsight, and we report it because the count is
the point: \textbf{thirteen instances in one project, and not one of them was
caught by a check written to catch it}. Eleven were caught by accident. One is a
narrow exception rather than a real one: a compiler refused the document over a
fragile macro in a caption, and the contradiction between that caption's two
sentences was then found by a human reading what the failure had put in front of
them. The compiler was checking the macro, not the claim, and the claim is what
was wrong. The thirteenth was caught by neither: it was found by putting a page
beside a page from another paper and seeing that ours looked softer. No check we
could plausibly have written would have produced that observation, and we record
it because a project that measures everything it can still needs someone to look
at the artefact. What separates the instances we now
catch automatically from those we do not is unglamorous: a test that fails when
the recorded value and the consumed value disagree. A field with no such test is
a comment that happens to live in a data structure, and an instrument with no
test of its own is a comment that happens to print.

\subsection{Why the two silent failure modes matter to anyone deploying a
low-resource clinical model}

Both failures in Section~\ref{sec:res:silent} share a property that makes them
dangerous in exactly the settings this kind of system is built for: every signal
an operator would normally check stays healthy.

\paragraph{A mis-shipped tokenizer is a packaging failure, not a modelling one.}
A model directory is copied, renamed, repacked and moved far more often than it
is retrained, and the tokenizer sits beside the weights rather than inside them.
When it goes missing the loader does not raise: it returns a vocabulary of five
tokens and the model answers its class prior on input it cannot read, with
well-formed probabilities, deterministic outputs and a correct label order. In a
clinic that failure looks like a working service. We mistook it for a collapsed
model in our own harness, and we caught it only because it contradicted a
confusion matrix we had recorded earlier: we caught it by
having an older measurement to disagree with, not by inspection. Any deployment
of a model whose inputs are in a script or language the operators cannot read
should verify the encoding at start-up: the vocabulary size and the
unknown-token rate on a fixed probe string cost nothing to check and are
decisive.

\paragraph{Surface fragility is cheap to measure and the measurement does not
identify its cause.} A corpus generated to a template has uniform case,
punctuation and spacing, and contains no typographical noise. A model trained on
it can change its answer for a third of inputs on capitalisation alone, and no
metric computed on that same corpus will show it, because the evaluation rows are
as uniform as the training rows. The check is cheap (perturb the surface and
count class changes), which is reason enough to run it. What it does not do is
locate the cause. As Section~\ref{sec:res:silent} sets out, a cased tokenizer
and an artefact whose confidence never left the region of its own decision
boundary account for the same flip rates as well as the corpus does, and we
measured nothing that separates them; the variant our tokenizer normalises away
changed nothing, which is a fact about the tokenizer rather than evidence about
the corpus. For low-resource clinical text, where patients type on phone
keyboards in a language with agglutinative morphology and no autocorrect,
surface variation is not an edge case but the ordinary condition of the input,
which is why the flip rate is worth reporting even while its cause is open.

\paragraph{Neither is visible from the model's own outputs.} This is the common
shape, and it is why we report both: a system can be wrong in a way that no
confidence score, no probability distribution and no accuracy figure reveals.
What detected each was a property check against something outside the model:
the encoding of its own input, and the invariance of its answer under a
transformation that should not matter.

%
%

\section{Limitations}
\label{sec:limitations}

\subsection{The three that bound everything else}

\paragraph{1. The evaluation set is nine sentences.} Rows built from one source
sentence are frame permutations of each other, not independent observations, so
the quantity governing every interval is the number of distinct sentences.
Section~\ref{sec:limits:base} gives the counts, generated from the frozen split.
Our own gate, which counts distinct sentences, refuses 38 of its cells on this set (a \emph{cell} is one language crossed with one urgency class, the unit the gate requires a minimum count of before it will report)
and reports nothing; the sizes it requires are derived rather than asserted:
710 distinct sentences for overall accuracy, 365 gold critical items per pure
language for critical recall, 720 for the critical-to-routine rate.
\textbf{No verdict on model quality is possible from this work, and none is
offered.} Every figure we report is a property of a corpus, of a specification, or
of an artefact's behaviour under perturbation, never of a model's clinical
performance.

\paragraph{2. No validated instrument authorises this construct.} We could
locate no instrument validated for the task this system performs, assigning
urgency from a patient's own written free text without examination, and
Section~\ref{sec:related:instruments} sets out what we did find and why none of
it supplies a reference standard. \textbf{Every urgency label in this work is
therefore an unvalidated construct.} No clinician has ratified any category definition, any threshold, or
any label in the corpus. The taxonomy names three classes; whether those classes
are the right ones, whether they can be assigned from text at all, and what each
would mean if they could, are open questions we are not qualified to settle.

\paragraph{3. One author, no clinical collaborator, one language.} This is a
single-author software project. No clinician has reviewed the corpus, the
taxonomy, or any output. No community health worker has been consulted about how
patients present. The corpus is Kinyarwanda only, and no second speaker has
reviewed it, so inter-rater agreement on anything (labels, naturalness,
register) is unmeasured because there is no second rater.

\subsection{How many distinct sentences each figure rests on}
\label{sec:limits:base}

The counts below are regenerated from the frozen split by the emitter, so they
cannot drift from the data they describe.
%
\begingroup
\renewcommand{\subsection}[1]{}%
\renewcommand{\label}[1]{}%
%
%
\subsection{Limitations}
\label{sec:limitations}

\paragraph{The reporting set is nine sentences, not eighteen thousand rows.} The held-out evaluation reports 17,942 rows, but those rows are frame permutations of only 9 distinct phrases in 5 phrase groups, and the row count is therefore not an independent sample size. Per class the position is starker:

\begin{center}
\begin{tabular}{lrr}
\toprule
Class & Distinct sentences & One sentence is \\
\midrule
CRITICAL & 4 & 0.25 of recall \\
URGENT & 4 & 0.25 of recall \\
ROUTINE & 1 & 1.00 of recall \\
\bottomrule
\end{tabular}
\end{center}

CRITICAL and URGENT recall therefore move in steps of approximately one quarter: a single sentence being read systematically as the neighbouring class costs about 25 percentage points, and no choice of optimiser or learning rate can recover it. ROUTINE is a single sentence, so its reported recall of 1.0 means that one sentence was classified correctly and should not be read as a class-level result. \textbf{A CRITICAL recall target of 0.95 over four sentences requires essentially all frames of all four to be correct}, and the distance between our result and that target is smaller than the granularity of the measurement.

\paragraph{Consequences for the sweep.} Three configurations spanning a threefold range of trainable parameters and a $1.5\times$ range of learning rate all left the same residual error, on the CRITICAL/URGENT boundary, while separating ROUTINE perfectly. We stopped searching on the grounds that the remaining error is a property of the corpus and the split rather than of the optimisation, and we report the configuration search as inconclusive on that boundary rather than as a tuning result.

\endgroup

\subsection{The paper's critique applies to the paper's own specification}

We argue that a number in a specification needs a stated source and a derivation
of what would make it measurable or achievable. \textbf{Our own specification
fails that test on every number it sets.} Each of the following was written
without a stated source, without a derivation, and without a check that it means
anything:

\begin{itemize}
\item critical recall $\geq 0.91$ in each pure language;
\item critical-to-routine rate $< 1.0\%$;
\item inference latency p50 $< 150$\,ms and p95 $< 200$\,ms (\emph{two} numbers, and the reason the count below is six across five bullets);
\item a cost ratio of 10:1 for a missed critical against any other error, which
  is the quantity that selects the served decision thresholds;
\item the requirement of 1{,}000{,}000 examples, which is the case this paper
  examines in detail.
\end{itemize}

We report this as one systemic failure with six instances across five bullets,
the latency requirement carrying two, rather than six
unrelated slips, and we do not exempt ourselves from it. The first four were
inherited from a requirements document written before any measurement; the fifth
is an engineering default we chose; the sixth we met in 130 seconds while failing
the quality gates written beside it. A reader is entitled to ask why we trust the
gates if we do not trust the requirements, and the next subsection is that answer.

\subsection{The gates are engineering defaults, not validated instruments}

The nine corpus gates are our own construction. Two have a derivation we would
defend (C1's cap on rows per seed and C2's floor on distinct seeds follow from
the cluster structure that governs an interval), and the rest are defaults
chosen by judgement. \textbf{C9's thresholds are placeholders}: the shares of
capitalised, unpunctuated, mis-spaced and mistyped items it requires were set by
us, not measured, and they should be replaced by the distribution a real pilot
exhibits once one exists. The near-duplicate threshold, the author-concentration
cap and the row cap are likewise conventions.

We state this rather than defend it. A gate whose number has no derivation is
open to exactly the criticism we level at the requirements, and the honest
position is that our instrument is a better-documented version of the same
species of artefact, not a validated one. What we would claim for it is narrower:
the gates make the corpus's properties visible and falsifiable, and the finding
that survives whatever thresholds are chosen is structural: the binding
quantity is authored seeds, and no row count substitutes for it.

\subsection{One generator, one language, one project}

Every measurement here comes from one generator, one language and one codebase.
The seeds-not-rows argument is a claim about a class of corpora (templated,
augmented or model-expanded), and we demonstrate it on a single member of that
class. We argue the generalisation from the structure of the cost asymmetry:
expansion is cheap and authoring is not, in any such pipeline. \textbf{We do not
demonstrate it}, and a reader should treat the general form as an argument to be
checked rather than a result we have established.

\paragraph{The experiment that would settle it, and what we predict it shows.}
The claim is testable on one generator and we state the design so that it can be
run against us. Hold the total row count fixed at roughly 30{,}000 and vary the
seed inventory across 10, 20, 40, 80 and 120 seeds, adjusting rows per seed to
keep the total constant. Evaluate every arm on one fixed held-out seed set,
disjoint from every training arm, with three runs per point at different seeds
and cluster-bootstrap intervals that resample the source sentence rather than the
row.

\textbf{We predict that held-out performance tracks the distinct-seed count and
is approximately flat in rows per seed}: the 10-seed arm and the 120-seed arm
differ substantially, while two arms with equal seeds and different expansion
factors do not separate beyond their intervals. That is the seeds-not-rows claim
stated as a falsifiable shape rather than as an argument from cost.

\paragraph{What it costs, measured rather than estimated.} We attempted the
sweep and did not finish it, and the cost is worth reporting because it names
what this experiment takes on the hardware the rest of the paper assumes. On a
four-core laptop with 1.7\,GiB of memory available, a single arm ran at
\textbf{11.1 seconds per optimisation step} with four threads and 9.5 with two:
doubling the threads made it about 17\% \emph{slower}, which we read as memory
contention rather than any property of the model. At that rate one arm of 940
steps is \textbf{174 minutes} and the fifteen-arm design is \textbf{roughly 43
hours} of uninterrupted compute.

The arms we built hold 3{,}000 rows rather than the 30{,}000 above, and the
figures are for that smaller budget. The reason is a property of the generator
worth stating on its own: \textbf{its output per seed varies twenty-one-fold},
from 386 rows for the smallest phrase to 8{,}129 for the largest. A fixed
30{,}000-row budget therefore selects for large seeds, since only 51 of 150
phrases can supply the 3{,}000 rows a ten-seed arm would need, and those 51 are
not label-balanced. Holding the budget at 3{,}000 is the largest round total at
which every arm can still draw seeds uniformly from the whole inventory, so the
arms differ in seed count and in nothing else. We reached step 600 of 940 on the first arm
before the machine restarted, and the harness has no resume within an arm, so
the attempt yielded no measurement.

We record this rather than quietly dropping the experiment. A design that is
affordable in principle and unaffordable on the machine available is the same
constraint this paper is about, one step further out: the sweep is cheap for
anyone with a cluster and out of reach for the setting the system is built for.
The arms, their manifests and the tests that hold them to one evaluation file
are committed and ready to run elsewhere.

\textbf{A flat curve would be a real result and we would report it as one.}
Saturation at a low seed count on a three-class task would mean the binding
quantity is smaller than we claim, and that the gate's floor of 3{,}000 distinct
seeds is set far above what this task needs. Either outcome is informative; only
declining to run it is not. The cost is roughly fifteen training runs at the
scale of the three already reported, which took 115, 70 and 95 minutes on two
CPU cores, so the whole sweep is on the order of a day of commodity compute and
needs no accelerator.

\paragraph{Figures that cannot be re-run}

The Phase~0 audit measured model accuracy, per-class recall, phrase-cluster
confidence intervals, calibration error and latency with scripts that were never
committed and no longer exist. \textbf{Those figures are NOT REPRODUCIBLE} and we
do not restate them in this paper, including the confidence intervals that would
otherwise make the nine-sentence point most vividly. Where an argument needed
them, we replaced them with something the repository can re-derive: the gate's
refusal, the power derivation, and the majority-class floors. The affected
sections of our own audit report carry the same mark.

\subsection{Further limitations}

\paragraph{One speaker, and no second-speaker review.} The Kinyarwanda
phrasings are one person's judgement. A second-speaker protocol is designed and
documented (a per-phrase rating scale to catch phrases that are wrong, and an
independent re-authoring pass to catch phrases that are merely not what someone
else would say), but it has not been run, and no ratings exist in the
repository. What one speaker can establish is that the phrasings are
consistently theirs; what makes them defensible is a second speaker agreeing
independently, and that has not happened. Where the speaker was uncertain they
consulted contacts informally, and one vocabulary question was still outstanding
with those contacts at the time of writing.

\paragraph{Provenance is mixed, and the corpus is not wholly speaker-authored.}
Roughly half the phrases are authored directly by the speaker and the remainder
are machine drafts under that speaker's approval, recorded per phrase in the
repository. We give no percentage: the per-phrase provenance is recorded, but no
committed script prints the split, so any figure here would be \textbf{NOT
REPRODUCIBLE}. A flag records an open clinical question, not a withdrawal, so a flagged phrase
may still be in the corpus. The three flag counts this paper uses are in
different units and are reconciled once, in
Section~\ref{sec:method:flagcounts}.

\paragraph{No clinician has signed off on anything.} \CorpusFlagged{} rows are
flagged as needing clinical review and \CorpusHeld{} are held pending decisions;
\CorpusFlaggedAndHeld{} are both.
The open questions are not cosmetic: whether fetal demise belongs in a triage
taxonomy at all, whether a patient reporting their own new confusion is
meaningfully confused, whether high fever with refusal to eat is urgent or
critical, and which of two descriptions of lower chest indrawing is the right
one. A clinician session is designed and has not been held. Two of the three
thresholds in our acceptance gate also await a clinical judgement that has not
been made; see Section~\ref{sec:gate}.

\paragraph{Further limitations are in Appendix~\ref{app:limits}.} Monolingual
scope, the converged holdouts, the absent baseline comparison, the six claims that
require studies nobody has run, the patient-facing text gap, the connectivity
assumption and the inconclusive configuration search are all recorded there.

%

\section{What the apparatus makes possible, and what it cannot settle}
\label{sec:future_work}

Nothing in this section is a plan with a date, and none of it asserts that the
system will work. The apparatus is built and the corpus is not; what follows is
what becomes measurable when authored seeds and a clinical lead exist, and what
remains open even then.

\subsection{What becomes possible once seeds exist}

The instruments in this paper are all written and tested against synthetic
inputs. Each has a first real use that is currently blocked on the same scarce
quantity.

\paragraph{The gates become a running constraint rather than a post-hoc audit.}
The authoring sheet issued to an author already carries the fields the gates
read: seed, origin, author, validator, reporter, age group, and the surface
variation the corpus lacks. Checking a returned sheet is seconds of compute, so
an author learns that a cell has become too uniform, or that one author now holds
too much of it, while there is still time to write differently. The gates were
written against a finished corpus; their useful position is at the point of
authoring.

\paragraph{The evaluation-set specification becomes a quota sheet.} The power
derivation fixes how many distinct sentences each gate cell needs. Those numbers
are an authoring budget once someone is authoring: they say how many independent
sentences per language and class must exist before any verdict is possible, and
they are the only honest basis we know of for costing the work.

\paragraph{The pipeline runs, or refuses, for a reason that can be acted on.} It
refuses today because the evaluation set is below its minimum. With a calibration
split and a test split that meet the specification, the same refusal becomes a
pass, and the artefacts it writes (the run manifest, the calibrated
temperature, the safety-tuned thresholds, the predictions carrying the served
decision) are the record a gate verdict would rest on.

\paragraph{The probe acquires a comparison.} Our surface-invariance figures are a
baseline taken on a model trained without surface variation. A model trained on a
corpus that carries it should flip less often, and that comparison is a real
result about the corpus rather than about either model. It is also the cheapest
check we know of that authored diversity had the effect it was supposed to have.

\subsection{Open questions we cannot answer from this project}

These are not tasks awaiting time. They are questions we are not positioned to
settle, stated so that someone who is may take them up.

\paragraph{What is ground truth for urgency assigned from patient-authored
text?} Every instrument we examined defines its categories on signs a clinician
observes in a present patient, and rules explicitly against substituting a
relative's account for observation. A label assigned from a written description
by someone who has not been examined is therefore not the same construct, even
when it uses the same category names. Whether a defensible ground truth exists
for it (a clinician's judgement given the text alone, agreement between
several such judgements, or the eventual disposition of the patient) is a
clinical and methodological question, not an engineering one. Until it is
settled, a label in a corpus like ours records what an annotator would say, and
the gap between that and what a patient needed is unmeasured.

\paragraph{How should effective diversity be measured in a synthetic
low-resource corpus?} We count distinct authored seeds because it is the quantity
that resists expansion and can be counted without argument. It is a proxy. Two
seeds may describe the same presentation in different words; one seed may carry
several distinct clinical situations. A lexical diversity measure computed over
one row per seed is the direction we chose, with a floor defined relative to a
pilot that does not yet exist, and we can neither calibrate nor defend that floor
today. What the right measure is, and whether any single number can stand for
diversity in a corpus assembled this way, is open.

\paragraph{What does calibration mean when confidence never reaches the review
threshold?}\label{sec:future:calibration} Our service refers any prediction below a confidence threshold to a
clinician. The artefact we probed never exceeded 0.55 on the critical class, so
under that rule every case would be referred: the threshold is not separating
confident from unconfident predictions, it is refusing all of them. A calibration
procedure fitted on such a model produces a temperature that is technically
correct and practically inert, and a review threshold set above the model's whole
confidence range is indistinguishable from having no model. Whether calibration
should be reported at all in that regime, and what a useful review threshold
would be when the distribution is that compressed, we do not know.

\paragraph{The other three languages, as they stand}

None of the other three contributes a row to any result in this paper; their
status, and the one result the English and French arms did produce, are in
Appendix~\ref{app:arms}.

\paragraph{The diagnostic that costs one inference pass}

The next diagnostic is per-sentence rather than per-configuration and costs one
inference pass rather than another training run; it is described in
Appendix~\ref{app:diagnostic}.

%

\section{Conclusion}
\label{sec:conclusion}

We set out to build an urgency classifier for patient-voice Kinyarwanda and
report instead what we found while trying to build one honestly.

A corpus requirement written as a row count constrained the quantity a machine
produces for free and left unconstrained the quantity only a person can produce.
We met the one-million-example target in 130 seconds with a corpus that fails
four of the nine quality gates written beside it, and six where its rows can be
attributed to a source sentence. The gate that binds counts distinct authored
seed phrases: at our inventory of 165, the largest corpus passing every gate is
zero rows at any row count. Two distances follow, both measured in sentences and
neither in rows. Passing the seed-count gate needs 2{,}835 further authored
sentences; reaching the one million rows the requirement names needs 19{,}835,
because a separate gate caps any one seed at 50 rows.

An evaluation set of 17{,}942 rows built from nine distinct sentences supports no
verdict, and our gate refuses on it rather than reporting a number. A model
trained on a corpus of uniform surface form changed its predicted urgency for
31.5\% of inputs under capitalisation alone and 21.0\% under a single typo. We
report those rates as a measurement and not as an attribution: the corpus is a
plausible contributor, and so are a cased tokenizer and an artefact sitting
close to its own decision boundary throughout, and we ran no probe that
separates them. A model directory shipped
without its tokenizer loads without error and answers its class prior on input it
cannot read, with well-formed probabilities, deterministic predictions and a
correct label order; we mistook it for single-class collapse in our own harness
until it contradicted a previously recorded confusion matrix.

What exists is an apparatus, not a result: a generator, a corpus gate checker, a
power derivation that fixes what a held-out set must contain, a deployment gate
that refuses rather than reports when it cannot measure, a malfunction probe, a
training pipeline that refuses to train below its evaluation-set minimum, and a
service that fails closed, never reassures a patient, and refuses to start on a
decision rule it cannot apply exactly. No model has passed the gate. The trained
artefact is an audit artefact and not a baseline.

What remains is not engineering. It is sentences written by native speakers:
2{,}835 of them to pass the seed-count gate, or 19{,}835 to reach the one
million rows the requirement names. It is also a clinical lead to define and
ratify what the categories mean, and a labelled evaluation set large enough in
distinct sentences for a verdict to be possible. Until those exist, the honest output of this project is the apparatus
and the negative results it produced.

Every number in this paper re-derives from a clean clone of the repository with a
single command, except those marked \textbf{NOT REPRODUCIBLE} where they appear.
Those rest on artefacts that were never committed; we mark them rather than
quietly dropping them, because an unreproducible figure a reader can see is
worth more than a gap they cannot.

\section*{Acknowledgements}
\label{sec:ack}

The Kinyarwanda phrases this corpus rests on were authored and ruled on by a
native speaker whose judgement overrode corpus frequency, pattern matching and
the author's drafts wherever they conflicted. Their name is withheld because we
have not yet obtained their explicit consent to be identified in a public
preprint. Consent to be named is not implied by having contributed, and this
paper will not assume it. The acknowledgement is owed and unpaid rather than
declined, and it is recorded here so that it stays visible as a debt.

No clinician contributed to this work. Nobody has reviewed the taxonomy, ratified
a category definition, or signed off a label; the clinician review pack exists
and has not been used.

\section*{Ethics statement}
\label{sec:ethics}

\textbf{No patient data of any kind was used in this work.} No real, identifiable
or de-identified patient record was collected, processed or stored at any point.
Every row of every corpus described here is synthetic: generated from
speaker-authored seed phrases, with no provenance in any clinical encounter.

Because no human-subjects data was involved, no ethical review was sought and
none was required. \textbf{This is a statement about what was not done, not a
clearance for what comes next.} The prospective work this paper points toward (a
nurse-baseline study, a clinician review session, community health worker
consultation, and any deployment in a health centre) involves human
participants and in some cases patients. Each requires approval from the Rwanda
National Health Research Committee or an equivalent institutional review board,
and compliance with Rwanda's data protection law, before it begins. None has
been sought, because none of those studies has been run.

The system described here has never been used with a patient, has never run in a
health centre, and is not cleared for clinical use. Its acceptance gate fails.

\section*{Data statement}
\label{sec:data}

\paragraph{What the corpus is.} A synthetic multilingual corpus for urgency
classification, generated from a small inventory of speaker-authored seed
phrases by a deterministic generator under a fixed seed. Its size is a
consequence of that inventory, not a target: the paper's central argument is
that a row count is the wrong unit and a seed count is the right one.

\paragraph{What is released, and what is not.} The generator, the split
manifests with their digests, the evaluation apparatus and every emitter that
produces a number in this paper are in the repository and re-derive from a clean
clone. The trained weights are not released: the model fails its own acceptance
gate, and publishing a checkpoint that a reader might deploy would be
irresponsible regardless of licence.

\paragraph{Licensed material is cited, not redistributed.} Clinical source
documents used as anchors, WHO ETAT among them, are under all-rights-reserved
terms that do not permit redistribution, including noncommercial redistribution.
They are cited by identifier and URL, and no such document is present in the
repository's working tree; the directory that held them is ignored, so none can
be restored by accident. We state the history exactly rather than describe the
repository as clean: the ETAT participant manual was committed in error on
2026-09-17 and removed on 2026-09-18, and because removing a file does not
remove it from version history, that blob is still reachable from earlier
commits. Purging it would require rewriting published history, breaking every
existing clone and every reference to a commit identifier, and we judged that
cost higher than the exposure of one WHO document that anyone may download from
WHO. It is therefore a recorded decision rather than an outstanding task. See
\path{docs/clinical/README.md}.

\paragraph{Language and provenance.} Every phrase carries a per-row provenance
record naming its origin and whether a speaker verified it. No Kinyarwanda in
this corpus was produced by machine translation from English. The other three
language arms are machine reviewed throughout and no native speaker of
English, French or Kiswahili has reviewed them; the paper does not claim
otherwise.

\bibliography{references}

\appendix
\newcommand{\sectionprefix}{Appendix~}
\newcommand{\sectionsuffix}{:}

\section{Person rulings, concept collapse and relation restrictions}
\label{app:person}

The authoring record holds \CorpusRows{} rows. A row is one concept crossed with
one grammatical person: first person (the patient speaking) or third person
(someone speaking about them, with the relation carried by a placeholder). Not
every concept supports both, and \CorpusInapplicable{} rows are ruled
inapplicable.

\textbf{This does not license dividing the row count by two to obtain a concept
total.} Section~\ref{sec:method:concepts} records that our own audit found the
concept count stated with five different values across this repository, and
declines to assert one; deriving 128 here would be asserting one of the five by
arithmetic instead of by claim. The rulings below are therefore counted in rows,
which is what the record enumerates.

The rulings fall into three groups, and each was made by a named party rather
than by default.

\paragraph{Whole concepts collapsed into another (15 concepts, 30 rows).} Where
two concepts turned out to name one sign, the speaker ruled them one concept and
the absorbed identifier went out of generation. The collapses are listed in
Table~\ref{tab:collapses}.

\begin{table*}[t]
\centering\small
\setlength{\tabcolsep}{4pt}
\begin{tabular}{*{6}{>{\raggedright\arraybackslash}p{1.6cm}}}
\toprule
Absorbed & Into & Absorbed & Into & Absorbed & Into \\
\midrule
EX17 & EX16 & HT06 & EX22 & NE04 & EX35 \\
EX30 & CR07 & IF07 & EX29 & NE08 & EX36 \\
EX42 & IF05 & NE01 & EX33 & PA01 & EX33 \\
GI08 & EX16 & NE02 & EX32 & PA06 & IF05 \\
HT01 & EX18 & NE03 & EX34 & PA10 & EX46 \\
\bottomrule
\end{tabular}
\caption{Concepts collapsed into another, 15 concepts over 30 rows.}
\label{tab:collapses}
\end{table*}

A collapse was permitted only where the
absorbing \emph{phrase} carried the whole sign; two proposed collapses were
rejected on that test, one because the absorbing phrase conditioned a general
danger sign on a temperature threshold and one because it covered only half of a
disjunctive sign.

\paragraph{First person removed (11 concepts).} Two grounds, both the speaker's.
The first is capacity: a patient who is unconscious or convulsing cannot report
it, which removes \texttt{EX32}, \texttt{EX33}, \texttt{PA02}, \texttt{PA03} and
\texttt{PA04}. The second is that the sign is an examination finding rather than
something the patient perceives (a skin pinch, sunken eyes, a growth chart)
which removes \texttt{GI04}, \texttt{PA05} and \texttt{PA07}. A separate standing
rule removes a paediatric first person that would merely duplicate an adult
concept, since the patient's being a child lives only in the relation word and
the first person has no slot for it; that removes \texttt{EX40}, \texttt{EX41}
and \texttt{EX43}.

\paragraph{Third person removed (9 concepts).} Nobody presents on another
person's behalf to collect a repeat prescription, attend their own routine review
or request their own screening test, so \texttt{CC08}, \texttt{CC09},
\texttt{CC10}, \texttt{EX10}, \texttt{EX11}, \texttt{EX44}, \texttt{EX45},
\texttt{PR06} and \texttt{PR07} have no third person.

\paragraph{Relation restriction.} Where a third person exists, the set of
relations it may be about is ruled per concept rather than defaulted: nine
concepts take no relations, seven are restricted to children, five to adults,
three to household members and one to the obstetric set. One concept is held
pending a decision and one is out of generation pending a service-design question
about whether men present for family-planning advice in this setting.

\paragraph{Open at the time of writing.} Of the \CorpusRows{} rows in the
authoring record, \CorpusFlagged{} carry a flag for clinician review and
\CorpusHeld{} are held pending a decision. \textbf{These two sets overlap:
\CorpusFlaggedAndHeld{} rows are both flagged and held}, \CorpusHeldOnly{} are
held without a flag, and \CorpusFlaggedOnly{} are flagged without a hold.
Subtracting the two counts from the row total therefore double-counts the
overlap.

\paragraph{Reconciling the record to the generating inventory.} A reader
subtracting the inapplicable and held rows from the row total arrives at
\CorpusRows{} $-$ \CorpusInapplicable{} $-$ \CorpusHeld{} $=$ 183, which is not
the \CorpusInventory{} phrases the generator holds, and an earlier version of
this appendix attributed the difference to the flagged/held overlap above. That
was wrong: \emph{inapplicable} and \emph{held} are disjoint, no row carries
both, and the overlap has no bearing on this subtraction. The subtraction is
short one term. \CorpusUnauthored{} rows are applicable and unheld but carry no
phrase yet, so they are in scope and not yet written, and \CorpusSecondPhrasings{}
row contributes a second phrasing, which is an additional corpus phrase rather
than an additional row. The reconciliation is therefore
\[ \CorpusRows{} - \CorpusInapplicable{} - \CorpusHeld{} - \CorpusUnauthored{}
   + \CorpusSecondPhrasings{} = \CorpusInventory{}, \]
every term of which is emitted from the record by
\path{review/emit_corpus_counts.py}. That emitter also reads
\path{dataset/vocabulary.py} and refuses to write if the two disagree, so the
figure the paper prints and the inventory the generator actually holds cannot
drift apart without the build saying so. The distinction the subtraction turns
on is between a row that may not be written and a row that nobody has written
yet, and only the second is work remaining.

Held and flagged are different states and the distinction is maintained
deliberately. A \emph{held} row generates nothing until the hold is lifted. A
\emph{flagged} row may still be in the corpus: the flag records that a clinical
question about it is open, not that it is withdrawn. That is why the two counts
above overlap rather than partition, and why \CorpusHeldOnly{} rows are held
without a flag while \CorpusFlaggedOnly{} are flagged without a hold.

The flags are themselves of two kinds: some are clinical questions about the
concept, which any language inherits, and some are doubts about a specific
Kinyarwanda word, which no other language inherits.

\section{Code-switching, pair by pair}
\label{app:codeswitch}

The corpus design names six ordered language pairs. \textbf{None of them produces
a row today}, and we state the position precisely because a corpus labelled
bilingual that is not is worse than one that admits the gap.

A Kinyarwanda speaker ruled fifteen medical terms for insertion, giving for each
whether it is genuinely switched or simply borrowed, its noun class, and the
agreement it triggers. That yielded five borrowed terms with established
Kinyarwanda forms, which belong in \emph{monolingual} rows rather than mixed
ones, and four cleanly switched terms. Table~\ref{tab:codeswitch} gives the
pair-by-pair position.

\begin{table*}[t]
\centering\small
\setlength{\tabcolsep}{4pt}
\begin{tabular}{>{\raggedright\arraybackslash}p{3cm}>{\raggedright\arraybackslash}p{11cm}}
\toprule
pair & why it does not generate \\
\midrule
kw $\rightarrow$ en & four switched terms are ruled, all nouns. Enough to emit
rows, not enough to be the mixed language the label claims. Pending a speaker's
answer on ten further terms. \\
kw $\rightarrow$ fr & no French-origin term is cleanly switched; all are
borrowed or dual-disposition. \\
en $\rightarrow$ kw, fr $\rightarrow$ kw & require the reverse insertion
direction and English or French frame fragments, neither of which exists. \\
sw $\rightarrow$ en, en $\rightarrow$ sw & require a Swahili-matrix worksheet.
None exists; the noun classes ruled so far are Kinyarwanda's. \\
\bottomrule
\end{tabular}
\caption{The six ordered language pairs, and why none of them generates a row.}
\label{tab:codeswitch}
\end{table*}

Five of the fifteen terms carry a dual disposition: the speaker's answer was
that they are both switched and borrowed, which is a real linguistic answer
rather than an incomplete one, but the two route a term to different places.
Two carry two candidate noun classes, and for one of those the two competing
forms sit in \emph{different} classes, so the choice is between two grammars
rather than two spellings.

\paragraph{Why the generator refuses rather than approximating.} The earlier
corpus produced mixed rows by swapping a whole phrase at one syntactic seam,
which accounted for nearly half its rows (recorded in
\texttt{docs/language-resources.md}; no committed script aggregates it) and added
no distinct clinical content. The
model that replaces it inserts a single content morpheme into an intact matrix
utterance and integrates it morphologically. That requires a noun class, and a
noun class is a fact about a language that cannot be inferred from the corpus.
The generator therefore declines to emit anything it cannot integrate, and
declines a pair whose inventory is too thin to be what the label claims.

Building the generator surfaced two errors it then prevented: an early version
filtered only on the embedded language and would have inserted
Kinyarwanda-classed terms into a Swahili frame, and a later one would have
shipped a four-noun inventory as a bilingual corpus.

\section{Leakage: closure, and the declaration that replaced it}
\label{app:leakage}

Our phrase holdout closes over nesting: where one phrase is contained in
another, or is an ordered subsequence of it, the two are inseparable. That
closure is computed from the strings and needs no outside knowledge. It is also
not sufficient.

A concept in this corpus has two phrasings, one in the patient's voice and one
in a carer's. They describe the same presentation and carry the same label, so
a holdout that puts one in training and the other in evaluation is not measuring
generalisation to an unseen presentation. Measured on the authored inventory,
\textbf{60 of the 61 concepts with both persons authored had them in separate
phrase groups}, a count recorded in \texttt{docs/phrase-group-closure.md} when
the closure was designed. No committed script recomputes it, so the figure is
\textbf{NOT REPRODUCIBLE} as it stands; the mechanism it demonstrates does not
depend on the exact count.

\paragraph{No similarity rule could have caught this, and the reason is
structural.} A third-person phrase begins with the relation placeholder and a
first-person phrase begins with a letter, so wherever the placeholder leads the
phrase the two share a prefix of zero characters: not a small overlap that a
lower threshold would reach, but none, however low the threshold goes.
Containment fails too, because the verb morphology differs between the persons
rather than one string embedding the other. The single concept that escaped is
the one that proves the mechanism: it places the placeholder mid-phrase, and its
two persons share 42 characters.

The fix is a declaration rather than a measurement. The authoring brief knows
which phrases belong to which concept, so the splitter reads that mapping and
unions every phrase of a concept into one group. Two properties of how it is
implemented matter more than the union itself. First, a declaration naming a
phrase that is not in the inventory is a fatal error rather than a no-op;
silence there would leave the concept's other phrasings unjoined and reopen the
leak with no signal. Second, the cost is paid openly: unioning concepts reduces
the number of independent evaluation units, and that reduction is the price of
the guarantee rather than an accident of it.

\paragraph{The uncomfortable part.} This is a leakage guarantee that no amount
of analysis of the corpus text could have produced. It depends on metadata
about who is speaking, which exists only because the authoring process recorded
it. A pipeline that received the same phrases as a flat list (which is how
such corpora are usually published) would compute a closure that looks
rigorous, report zero overlap honestly, and split 60 of 61 concepts across the
holdout. We had already made exactly this class of mistake once, in an earlier
split whose manifest recorded complete leakage in a field nobody read.

\paragraph{The same shape, three times.} The split manifest recorded complete
leakage in a machine-readable field, in the commit that created it, and the two
fields being read were both honestly zero. Our acceptance threshold existed as
independent copies of one constant in two source files, so changing the safety
threshold in one would silently not apply in the other.

The third is this paper's own tables. \path{training/holdout_eval.py} wrote
\path{results_table.tex}, \path{confusion_table.tex} and
\path{sweep_table.tex} on every evaluation pass, and until 2026-09-18 no file
included them. The build header at the top of \path{main.tex} had said since
the pipeline was written that \path{results.tex} inputs the tables. It did not,
and \textbf{nothing failed, because a generated file nobody reads produces no
error}: no missing input, no undefined macro, no warning. The emitter kept its
side of the contract and the document quietly stopped keeping its. They are
restored in Appendix~\ref{app:run}.

None of the three is a measurement failure. Each is a fact recorded where the
code path that needed it does not look, and a project accumulating rulings
faster than it accumulates places to read them should expect more of them. The
third is the one that should worry a reader most, because the other two were
caught by a person re-reading a file, while this one was caught only by counting
the tables in a rebuilt archive against the tables the emitter writes.

\section{The v1 family holdout, in full}
\label{app:v1leak}

\paragraph{One of the two evaluation splits never worked, and the manifest
recorded it from the first commit.} v1 was split two ways, by phrase and by
family, and presented as two difficulty levels. The family holdout was not a
holdout. Its own frozen manifest reports that every one of its 114{,}321
evaluation rows was built on a phrase that also appears in training
(Table~\ref{tab:v1manifest}).

\begin{table*}[t]
\centering\small
\setlength{\tabcolsep}{4pt}
\begin{tabular}{>{\raggedright\arraybackslash}p{8cm}r}
\toprule
\path{eval_rows_whose_phrase_appears_in_train} & 114{,}321 \\
\path{eval_rows_leaked_fraction} & 1.0 \\
\path{phrase_overlap} & 50 \\
\path{substring_violations} & 54 \\
\midrule
\path{exact_text_overlap} & 0 \\
\path{family_overlap} & 0 \\
\bottomrule
\end{tabular}
\caption{The v1 family holdout's own frozen manifest, as committed.}
\label{tab:v1manifest}
\end{table*}

The failure is not that the leakage went undetected. It was detected, measured
and written into the manifest by the same commit that introduced the pipeline.
It went unread. The two fields anyone looked at, exact text overlap and
family overlap, were both zero, and they were zero honestly: no evaluation
row appeared verbatim in training, and no family straddled the split. Holding
out a \emph{family} does not hold out a \emph{phrase}, and it was the phrase
that carried the label. A number that says the holdout is worthless can sit in
a machine-readable artefact for as long as nobody reads that field.

v1's phrase holdout, by contrast, was internally clean, as v2's is.

\paragraph{A separate contamination, in the other direction.} Scoring a
phrase-trained model on the family evaluation set was also invalid: 89.2\% of
family-evaluation rows (101{,}945 of 114{,}321) appeared verbatim in
phrase-split training. This is a \emph{cross}-split measurement and is not the
same failure as the one above; conflating the two is a mistake this project
made and corrected in its own records.

We state plainly that the cross-split figure is \emph{worse} in v2, not
better: it is 100.0\% in both directions (24{,}900 of 24{,}900 and 34{,}425 of
34{,}425), because the two v2 evaluation sets are disjoint partitions of one
corpus, so each trains on everything the other holds out. What v2 fixed is the
within-split failure: family-evaluation against family-training is now
0.0\%, where v1 was 100\%. The two splits swapped failure modes rather than
one improving, and a table placing a v1 family result beside a v2 one would
compare a wholly contaminated evaluation with an uncontaminated one and call
the difference a model improvement.

\section{The cost of the configuration search}
\label{app:search}

Three configurations were trained to completion, at 115, 70 and 95 minutes on
two CPU cores. The whole search cost under five hours of compute on a laptop. We
report no result from any of them: these are costs, recorded in the run records,
and the artefact whose behaviour Section~\ref{sec:res:silent} probes is one of
the three.

We state this deliberately. Work of this kind is often reported without cost,
which leaves a reader unable to tell an exhaustive sweep from three runs, and
leaves anyone without a cluster unsure whether the method is available to them.
It also bounds what the search establishes: three points do not characterise a
response surface, and we describe the outcome as inconclusive on the
CRITICAL/URGENT boundary rather than as a tuned result.

The affordability follows from the corpus rather than from any efficiency
technique, and it is the seeds-not-rows point from the training side: the task is
\textbf{150 distinct training phrases} mapped to three labels, however many rows
those phrases were expanded into. The model is a small multilingual encoder with
its whole embedding module frozen (the word-embedding table together with the
position and token-type tables and the embedding LayerNorm; see
Section~\ref{sec:method:training}), leaving between 3.7 and 10.8 million trainable
parameters. A budget of a few hundred optimisation steps is not a compromise
forced by the hardware but the right size for the problem, and the run that
preceded these was abandoned in part because it had been budgeted at roughly
forty times what the task needed.

\section{The reported run in full}
\label{app:run}

These three tables were emitted on every run and included by no file until
2026-09-18; that is the third instance of the recorded-but-unread pattern and is
treated in Appendix~\ref{app:leakage}. They are collected here rather than in
the results section because none of them is offered as a statement about model
quality, for the reasons in Section~\ref{sec:res:none}; they are the evidence
behind claims the paper makes about its own \emph{gate}, and a reader checking
those claims should not have to re-run the pipeline to see the numbers.

Read them against that section and not as a performance report. In particular,
Table~\ref{tab:results} is measured on 9 distinct sentences, so its support
column is not an independent sample size, and Table~\ref{tab:confusion} gives
the confusion matrix those per-class figures are computed from.

%
%
\begin{table*}[t]
\centering
\small
\setlength{\tabcolsep}{4pt}
\caption{Triage performance of the reported model on the frozen phrase holdout. The 17,942 rows are frame permutations of 9 distinct phrases in 5 phrase groups; the support column is therefore not an independent sample size, and the final column gives the count that governs the granularity of each recall figure.}
\label{tab:results}
\begin{tabular}{lrrrrr}
\toprule
Class & Precision & Recall & F1 & Rows & Distinct sentences \\
\midrule
CRITICAL & 0.6633 & 0.8504 & 0.7453 & 8,962 & 4 \\
URGENT & 0.7426 & 0.4963 & 0.5950 & 7,793 & 4 \\
ROUTINE & 0.9549 & 1.0000 & 0.9770 & 1,187 & 1 \\
\midrule
Macro avg & n/a & n/a & 0.7724 & 17,942 & 9 \\
\bottomrule
\end{tabular}
\\[2pt]{\scriptsize Run fingerprint \texttt{a288e8ca47f375f6}}
\end{table*}

%
%
\begin{table*}[t]
\centering
\caption{Confusion matrix for model\_v2d\_freeze8\_lr1e-5 on the reporting set. ROUTINE is separated perfectly; the residual error is entirely on the CRITICAL/URGENT boundary, and this was true of every configuration trained.}
\label{tab:confusion}
\small
\setlength{\tabcolsep}{4pt}
\begin{tabular}{>{\raggedright\arraybackslash}p{4cm}rrr}
\toprule
Truth $\downarrow$ / Pred $\rightarrow$ & CRITICAL & URGENT & ROUTINE \\
\midrule
CRITICAL & 7,621 & 1,341 & 0 \\
URGENT & 3,869 & 3,868 & 56 \\
ROUTINE & 0 & 0 & 1,187 \\
\bottomrule
\end{tabular}
\\[2pt]{\scriptsize Run fingerprint \texttt{a288e8ca47f375f6}}
\end{table*}

%
%
\begin{table*}[t]
\centering
\caption{Three configurations on the same frozen phrase holdout, same split seed, same reporting set. Trainable parameters exclude the frozen embedding module, which is 96{,}199{,}296 of the model's 117{,}641{,}859 parameters in every row: the 250{,}002$\times$384 word-embedding table (96{,}000{,}768) together with the position and token-type tables and the embedding LayerNorm.}
\label{tab:sweep}
\small
\setlength{\tabcolsep}{4pt}
\begin{tabular}{>{\raggedright\arraybackslash}p{3.6cm}rrrrr}
\toprule
Run & Trainable & Macro F1 & CRIT P/R & URG P/R & Gate \\
\midrule
v2b\_freeze10\_lr1e-5 & 3,697,923 & 0.7660 & 0.7951/0.9007 & 0.8636/0.6315 & FAIL \\
v2c\_freeze6\_lr1.5e-5 & 10,795,779 & 0.5545 & 0.5337/0.9749 & 0.2321/0.0083 & FAIL \\
v2d\_freeze8\_lr1e-5 & 7,246,851 & 0.7724 & 0.6633/0.8504 & 0.7426/0.4963 & FAIL \\
\bottomrule
\end{tabular}
\\[2pt]{\scriptsize All rows evaluated in one pass; fingerprint of the reported model \texttt{a288e8ca47f375f6}}
\end{table*}

Table~\ref{tab:sweep} is the one that carries an argument. The three
configurations differ only in how many layers were frozen and in learning rate,
they were scored in one pass on the same reporting set, and all three fail the
acceptance gate. The middle row is the configuration of
Section~\ref{sec:gate-failure}: the highest CRITICAL recall of the three, an
URGENT recall of 0.0083, and the lowest macro F1. A gate reading CRITICAL
recall alone would have selected it.

\section{The other three language arms}
\label{app:arms}

The first part records what each arm contains today; the second records what
each would still need.

\subsection{As built}

The corpus reported here is Kinyarwanda. The remaining three arms exist at
three different stages, and the distinction matters because their provenance
differs by component rather than uniformly.

\paragraph{Relation terms.} A third-person phrase carries a placeholder that
expands to a set of relation words. For Swahili these are \emph{speaker
authored}: a Kiswahili speaker supplied all twelve with regional variants and
notes. For English and French they are not: their wording was taken from the
v1 corpus's machine-drafted subject slot. In every arm the set \emph{membership}
is the Kinyarwanda speaker's ruling mirrored one for one, so a Swahili
third-person row carries speaker-authored wording under a mirrored membership
decision, and the paper should not describe either half as the other.

One Kinyarwanda relation has no Swahili counterpart and the slot is left empty
rather than filled. Kinyarwanda's eighth relation denotes an elderly woman
generally rather than the speaker's own grandmother; the returned Swahili terms
are possessive and name a different referent, so they are recorded and not
wired into a set that means something else.

\paragraph{No rating session has run, so authenticity is unmeasured.} Whether
these phrases read as something a patient would say has not been measured. The instrument, a five-point scale whose middle band is
deliberately ``understandable, but not how I would put it'', the state the v1
corpus was in, is written and unexecuted.

\paragraph{Phrases.} No phrase is authored in any of the three. English and
French hold machine-drafted candidates on the same concept spine, unreviewed by
any speaker of either language. Swahili holds none: its authoring brief contains
no Swahili at all, deliberately, so that a fluent draft cannot anchor the
speaker to its errors: the finding that caused the Kinyarwanda arm to be
re-authored rather than corrected.

\paragraph{Frame fragments.} The openers, time expressions, contexts and
closing lines that wrap a phrase are authored for Kinyarwanda only. \textbf{No
other language can generate a single row}, whatever its phrase count, and this
rather than the phrase inventory is the binding constraint on a four-language
corpus.

\subsection{As they stand for future work}

The corpus is Kinyarwanda only, and none of the other three contributes a row to
any result in this paper.

\paragraph{English and French exist as review briefs, not corpora.} Both are
machine-drafted candidate phrasings with per-row provenance and review verdicts
against the same concept spine. No English speaker and no French speaker has
ruled on them, and the French brief records explicitly that no French speaker has
seen any of it. The eleven held rows both arms lifted, across seven concepts,
are treated as a finding in Section~\ref{sec:method:arms} rather than here,
including which of them were reached independently and which were mirrored.

\paragraph{Swahili is out for authoring, with no draft supplied.} The brief the
speaker received contains no Swahili at all: it carries the English gloss, the
person split, the applicability and relation rulings and the open clinical
questions, and an empty column. This follows the Kinyarwanda finding that a
fluent draft anchors a speaker to its errors, which is why that arm was
re-authored from scratch rather than corrected. A machine-translated Swahili
corpus exists in the frozen v1 vocabulary and is deliberately withheld until the
authored phrases exist, which is the only order in which comparing them means
anything. The reasoning is in Section~\ref{sec:method:arms}.

One gap is recorded and not closed: the frame slots have not been authored, so
the arm emits nothing regardless of what else it holds. \textbf{An earlier
version of this appendix also claimed that no Swahili relation terms exist
anywhere in the project. That was false, and it contradicted this appendix's own
opening paragraph.} The terms exist, speaker-authored with regional variants, in
\path{review/swahili_relations.py}; seven are wired and the eighth Kinyarwanda
relation has no counterpart, for the reason given above. What is true is that no
frame can use them yet. The two statements were written months apart and nothing
compared them, which is why Table~\ref{tab:arms} is emitted from the inventory
rather than described in prose.

\section{The per-sentence diagnostic}
\label{app:diagnostic}

Every configuration trained separated the routine class without error and none
resolved the critical/urgent boundary. The next diagnostic is per-sentence rather
than per-configuration: which held-out sentences are read as the adjacent class,
and whether they are the same sentences across configurations. That distinguishes
an ambiguous concept from a mislabelled one from an artefact of the split. It is
one inference pass rather than another training run, and it is blocked on nothing
except a set large enough for the answer to mean something.

\section{Further limitations}
\label{app:limits}

\paragraph{Monolingual.} The corpus is Kinyarwanda. English and French exist
only as unverified machine-drafted review briefs, Swahili is out for authoring,
and none of the three contributes a row to any result reported here. The system
is therefore not evaluated on the multilingual setting it is ultimately
intended for.

\paragraph{The two holdouts have converged, so there is only one evaluation.}
The corpus is split by phrase and by family, and it would be wrong to present
these as two difficulty levels. Because v2 is monolingual, each phrase belongs
to exactly one family, so holding out a family removes its phrases entirely: the
two splits are one strictness at two sampling ratios. Worse, they are disjoint
partitions of the same corpus, so scoring a model trained on one against the
other's evaluation set measures memorisation and nothing else: 100.0\% of
rows in both directions. Only one honest evaluation number exists per trained
model, and it is the one we report.

\paragraph{No model meets the acceptance gate, and none has been trained
through the pipeline that would produce a deployable one.} The artefact that
exists does not meet the gate; the pipeline that would calibrate a model and tune
its thresholds refuses before reading the training corpus, because the
evaluation set is below its minimum. Nothing here is in use with any patient.

\paragraph{One substantiation source sits in a repository that also ships
machine-generated material.} One resource we consulted distributes human-written
and model-generated content from the same repository, and our record of its
composition comes from its own dataset card. \textbf{We could not read that card
during verification}: the item returned an HTTP error to every automated
request, so we quote no counts from it and rely on it for nothing in this
paper.

\paragraph{No baseline comparison exists.} We did not train a comparison
model. The obvious baselines (multilingual BERT, AfriBERTa) were scoped
and not run, so we report no comparison table and make no claim about how this
model performs relative to alternatives. What we report is one model family
across three configurations. A reader wanting to know whether a different
pretrained encoder would do better on this corpus should read that question as
open, not as answered in our favour.

\paragraph{Six claims require studies that have not run.} No value for any of
the following exists anywhere in our records, and none is estimated here:

%
\begin{center}
\footnotesize
\setlength{\tabcolsep}{4pt}
\begin{tabular}{>{\raggedright\arraybackslash}p{2.4cm}>{\raggedright\arraybackslash}p{4.6cm}}
\toprule
Claim & Why there is no value \\
\midrule
Inter-rater agreement ($\kappa$) & There is no second annotator, so there is no
agreement statistic: not a low one, none \\
Taxonomy approved by a clinician & No clinician has reviewed it. We approached
clinical contacts and did not secure one \\
Native-speaker authenticity ratings & No rating session has been held. The
five-point instrument is written and unexecuted \\
Human-nurse baseline & No nurse study has been run, so we cannot say how a
person would triage the same text \\
Community health worker consultation & No consultation took place \\
Deployment & The system has never run in a health centre and is not cleared to \\
\bottomrule
\end{tabular}
\end{center}

The protocol for each is written and ready to execute; what is missing is a
person, not an instrument. None of these absences was chosen. Each is a study we
could not arrange, and we list them rather than omitting them because a claim
quietly dropped from a paper is indistinguishable, to a reader, from one that was
never made.

\paragraph{Patient-facing text exists in two of four languages.} The sentence a
patient reads, and the SMS they receive, is speaker-authored or absent,
and it is authored for Kinyarwanda and Swahili only. English and French return
an explicit pending state. The system runs; it declines to say anything to a
patient in a language no speaker has written for.

\paragraph{The client requires connectivity, in a setting that does not have
it.} The system is intended for health centres with intermittent connectivity,
and neither half of it tolerates an outage. Classification is server-side: a
449\,MB encoder that will not run in a browser on the hardware these clinics
have, so a disconnected client cannot triage at all; it could at best capture
symptoms for later. The queue view is worse than unavailable during an outage:
it retains the last rows it successfully fetched and continues to render them
without an age, so a board that is minutes stale reads as current, and a patient
who arrived during the outage is simply absent from it. Submissions made while
disconnected are lost rather than queued. We record this as a constraint rather
than reporting a solution: an offline mode for this system is a capture-and-sync
design needing server-side idempotency and a conflict rule for replayed triage,
and neither exists. The full inventory is in \texttt{docs/frontend-limitations.md}.

\paragraph{The configuration search is inconclusive rather than complete.}
Three configurations were trained and the search was stopped by judgement, not
by convergence. We did not tune to the gate, and we did not run the
per-sentence diagnostic that would distinguish an ambiguous concept from a
mislabelled one on the unresolved boundary.

\end{document}